\documentclass[letterpaper]{article} 
\usepackage{aaai25}  
\usepackage{times}  
\usepackage{helvet}  
\usepackage{courier}  
\usepackage[hyphens]{url}  
\usepackage{graphicx} 
\usepackage{natbib}  
\usepackage{caption} 
\usepackage{algorithm}
\usepackage{algorithmic}

\usepackage{newfloat}
\usepackage{listings}

\usepackage{graphicx}
\usepackage{subfigure} 
\usepackage{epstopdf}
\usepackage{colortbl}
\usepackage{multirow}
\usepackage{amsmath}
\usepackage{amssymb}
\usepackage{appendix}
\usepackage{bm}
\usepackage{xcolor}
\usepackage{booktabs}

\definecolor{myblue}{HTML}{BFEEF4}

\DeclareCaptionStyle{ruled}{labelfont=normalfont,labelsep=colon,strut=off} 
\floatstyle{ruled}
\newfloat{listing}{tb}{lst}{}
\floatname{listing}{Listing}
\title{Towards Robust Incremental Learning under Ambiguous Supervision}

\author{
    Rui Wang\textsuperscript{1,2},
    Mingxuan Xia\textsuperscript{1,2},
    Haobo Wang\textsuperscript{1,2},\\
    Lei Feng\textsuperscript{4} ,
    Junbo Zhao\textsuperscript{3},
    Gang Chen\textsuperscript{3},
    Chang Yao\textsuperscript{1,2*}
}
\affiliations{
    \textsuperscript{\rm 1}School of Software Technology, Zhejiang University\\
    \textsuperscript{\rm 2}Hangzhou High-Tech Zone (Binjiang) Institute of Blockchain and Data Security \\
    \textsuperscript{\rm 3}College of Computer Science and Technology, Zhejiang University \\
    \textsuperscript{\rm 4}School of Computer Science and Engineering, Southeast University, China \\
   \{r.wang, xiamingxuan, wanghaobo\}@zju.edu.cn, fenglei@seu.edu.cn, \{j.zhao, cg, changy\}@zju.edu.cn

}

\usepackage{bibentry}

\begin{document}

\maketitle

\begin{abstract}
Traditional Incremental Learning (IL) targets to handle sequential fully-supervised learning problems where novel classes emerge from time to time. However, due to inherent annotation uncertainty and ambiguity, collecting high-quality annotated data in a dynamic learning system can be extremely expensive. To mitigate this problem, we propose a novel weakly-supervised learning paradigm called Incremental Partial Label Learning (IPLL), where the sequentially arrived data relate to a set of candidate labels rather than the ground truth. Technically, we develop the Prototype-Guided Disambiguation and Replay Algorithm (PGDR) which leverages the class prototypes as a proxy to mitigate two intertwined challenges in IPLL, i.e., label ambiguity and catastrophic forgetting. To handle the former, PGDR encapsulates a momentum-based pseudo-labeling algorithm along with prototype-guided initialization, resulting in a balanced perception of classes. To alleviate forgetting, we develop a memory replay technique that collects well-disambiguated samples while maintaining representativeness and diversity. By jointly distilling knowledge from curated memory data, our framework exhibits a great disambiguation ability for samples of new tasks and achieves less forgetting of knowledge. Extensive experiments demonstrate that PGDR achieves superior performance over the baselines in the IPLL task.
\end{abstract}

\section{Introduction}
Modern deep models are mostly developed in curated and static benchmark datasets, but data in the real world typically emerge dynamically. This motivates the study of incremental learning (IL) \cite{ca2,rlip,KimPH24} that enables models to learn from sequentially arriving tasks. 
Despite the flexibility, it is well known that deep models struggle to retain their known concepts, i.e., \textit{catastrophic forgetting}, making it hard to gradually accumulate knowledge. To alleviate this problem, a plethora of IL methods have been studied, including replay-based methods~\cite{icarl,ancl,Zheng0YZ24}, regularization-based methods~\cite{SI,szatkowski2023adapt}, architecture-based methods~\cite{den,MaroufRTL24} and so on~\cite{chenshen2018memory,SeoKP23,MarczakTTC24}.


\begin{figure}[!t]
  \centering
  \includegraphics[height=2.5cm]{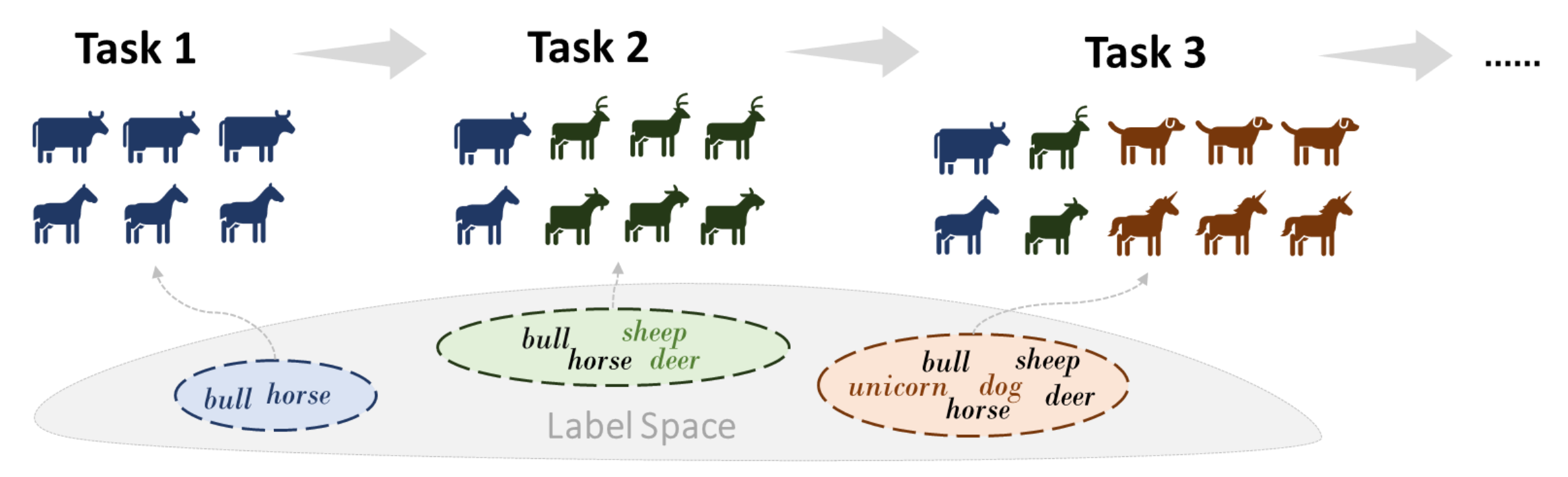}
   \caption{In the first task, all samples are from new classes, while subsequent tasks consist of samples containing both new and old classes. Each sample is assigned a set of candidate labels, ensuring the inclusion of the true label.}
   \label{fig:fig1}
\end{figure}

Traditional IL methods~\cite{icarl,wa} are built on the assumption that data is accurately annotated. However, real-world data often has label ambiguity, making precise annotations labor-intensive, especially in sequential learning. For instance, the Alaskan Malamute can be visually similar to the Siberian Husky, hindering non-experts from accurately identifying the true breeds.
Recently, this ambiguous supervision problem has attracted great attention from the community \cite{LyuWF22,abs_analyze,mlpll,JiaP0Z24}.

To reduce annotation costs, we study a novel weakly supervised learning framework dubbed \textit{incremental partial label learning} (IPLL), where (i)-data is given sequentially as a stream; (ii)-each task contains a vast number of new class samples while potentially carries old class data, and (iii)-each sample is associated with a candidate label set instead of the ground truth; see Figure~\ref{fig:fig1}. Notably, since the annotator may confuse previous experience with the true label of the sample, our IPLL setup allows the candidate label set of the sample to include both new and old classes, while ensuring the inclusion of the true one.
Arguably, the IPLL problem is deemed more practical in real-world scenarios due to its relatively lower cost to annotations.

\begin{figure}
	\centering
	\subfigure[Class prior estimation.]{
		\begin{minipage}[b]{0.47\textwidth}
			\includegraphics[width=1\textwidth]{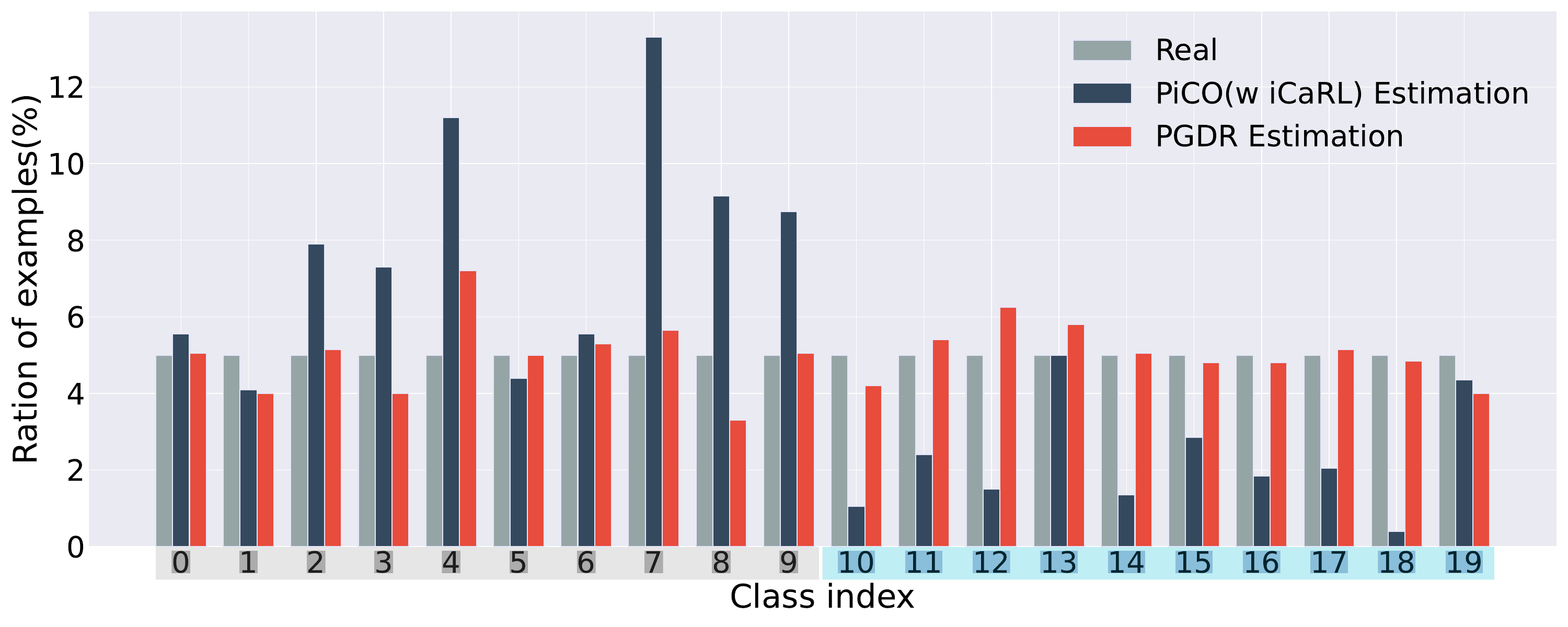}
		\end{minipage}
		\label{fig:estimation}
	}\\
        \centering
    	\subfigure[Per-Task accuracy.]{
    		\begin{minipage}[b]{0.38\textwidth}
   		 	\includegraphics[width=1\textwidth]{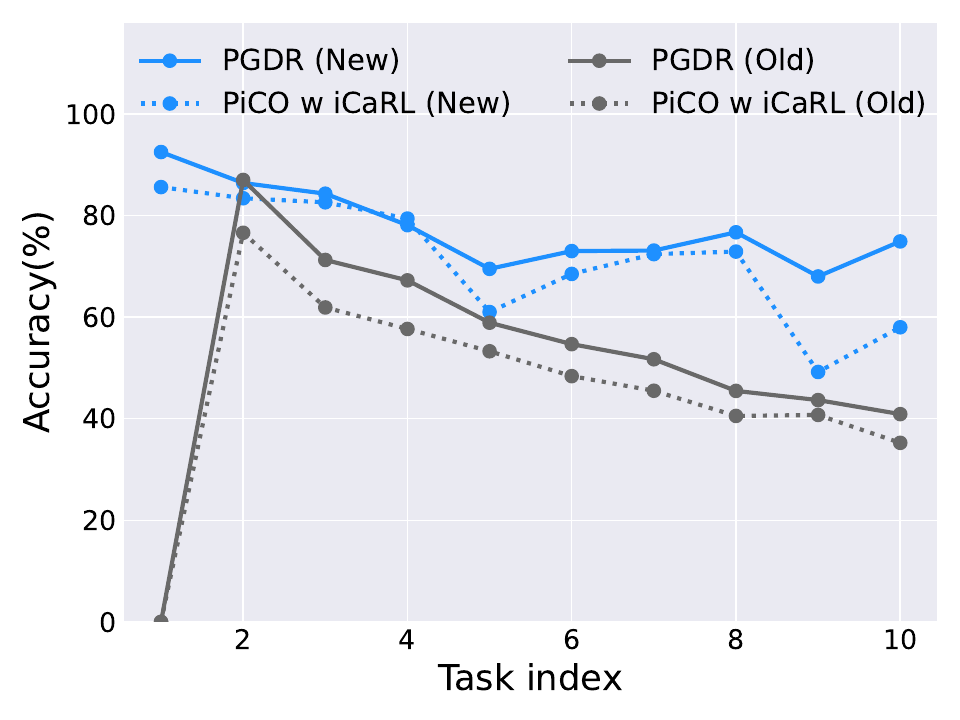}
    		\end{minipage}
		\label{fig:q_02_task1}
    	}
	\caption{(a) Class distribution based on both predicted labels and ground-truth labels. The real/estimated distribution of \colorbox{gray!30}{old classes} (0$\sim$9) and \colorbox{myblue}{new classes} (10$\sim$19) in the early stage of the second task of the CIFAR100 (10 tasks). Combining the PLL baseline (PiCO) and the IL baseline (iCaRL) leads to {\bf{classification bias}}. (b) Comparison of our PGDR and PiCO with iCaRL in IPLL. It displays the average accuracy of new and old classes for each task in CIFAR100 where PiCO with iCaRL demonstrates inferior and unstable performance compared to our method.}
 \vspace{-0.3cm}
\end{figure}

The key to successful learning from ambiguous supervision is label disambiguation, i.e., identifying the true labels from the candidate sets.
To achieve this, existing partial label learning (PLL) algorithms~\cite{pico,mlpll,LiJLWO23,JiaP0Z24,liulearning} mostly rely on the self-training algorithm that assigns pseudo-labels using the model predictions. However, such a strategy can be problematic in our IPLL framework.
On the one hand, the dynamically changing environment leads to the forgetting of past knowledge, which amplifies knowledge confusion that \textbf{hinders the disambiguation procedure}.
On the other hand, ambiguous supervision significantly undermines the ability of DNNs to acquire meaningful knowledge from samples, \textbf{exacerbating catastrophic forgetting}.
As exemplified in our empirical studies, directly combining the PLL method with the IL method also leads to inferior results. Specifically, when combining the most popular PLL method PiCO~\cite{pico} with one classic IL method iCaRL~\cite{icarl}, the model overly emphasizes old classes in the early stages of each task (see Figure~\ref{fig:estimation}), thus exacerbating the confusion of old class knowledge (see Figure~\ref{fig:q_02_task1}). In the more challenging Tiny-ImageNet, the negative impacts of classification bias become more pronounced (see Figure~\ref{fig:Disambiguation_old}).
 This gives rise to the fundamental issue in IPLL—\textit{how to balance the model's perception of new and old knowledge}.

To address this issue, we propose the {\bf{P}}rototype-{\bf{G}}uided {\bf{D}}isambiguation and {\bf{R}}eplay Algorithm (PGDR), which leverages class prototypes carrying rich category information as proxies to balance the model's perception of classes. Concretely, to alleviate label confusion, we propose a \textit{prototype-guided label disambiguation} strategy, which firstly performs distance-based old/new sample separation and then, momentumly updates on pseudo-labels. 
Secondly, to mitigate catastrophic forgetting, we construct an episodic memory by distance-based sampling for representative samples and region-aware sampling for diverse samples, which are replayed by a knowledge distillation loss.
These two modules mutually benefit each other to achieve a balanced perception on all classes---the disambiguation module ensures accurate training on new classes while the memory replay module strengthens the old classes.
We conduct comprehensive experiments on benchmark datasets to show that PGDR establishes state-of-the-art performance.
In IPLL, our method outperforms the best baseline by {\bf{6.05\%}} and {\bf{11.60\%}} on the CIFAR100 and Tiny-ImageNet.

\section{Related Work}
\paragraph{Incremental Learning. }\label{sec:subsec2}
The data appear in a sequence and the model continuously learns novel knowledge while maintaining the discrimination ability for previous knowledge. There is a classical incremental learning variant---blurry incremental learning~\cite{blurry,MoonPKP23}, where different stages exhibit distinct data distributions. Due to the intersection of the label space, it faces heightened ambiguity.
Nonetheless, IL and its variants are confronted with the challenge of knowledge forgetting \cite{KimPH24,WCRKZ24}. To address this issue, existing methods can be categorized into three major strategies.
For architecture-based methods~\cite{apd,MaroufRTL24}, different model parameters are allocated for each task.
The regularization-based methods~\cite{c-flat} introduce regularization terms into the objective function. 
The replay-based methods~\cite{tpami,Yoo0WP24} retain a subset of historical samples and incorporate them into subsequent tasks.
Moreover, while some works~\cite{LangeT21,AsadiDMAB23} employ prototypes, they mainly focus on using them to alleviate knowledge forgetting and do not investigate their effectiveness in other aspects, distinguishing our approach from theirs.

The IL variants discussed above all assume accurately labeled data. More and more researchers are increasingly interested in incremental learning in limited or unsupervised settings~\cite{WuCWF23}, such as unsupervised continual learning~\cite{ChaCM24}, semi-supervised continual learning~\cite{fan2024persistence}, and noisy labeled continual learning~\cite{noisy1}. 
However, most of the work overlooks label ambiguity. Although~\cite{DBLP:journals/ml/YuWZ24} attempts to address this issue, the subsequent samples are unlabeled, making it similar to a semi-supervised learning problem.

\paragraph{Partial-Label Learning.}\label{sec:subsec3}
The fundamental challenge in PLL is label disambiguation, requiring the model to select the true label from the set of candidate labels~\cite{YanG21,DBLP:journals/tkde/LyuFWLL21,DBLP:journals/pami/WangZL22,terial}. Some PLL methods enhance candidate label disambiguation through adversarial learning~\cite{zhang2020partial,ZhangLYZNCL23}, and the majority of work~\cite{pico,CroSel} is geared towards devising appropriate objectives for PLL. The strategy~\cite{abs_analyze} based on averaging assigns equal weights to the labels in the candidate labels, and then obtains predictions by averaging the output. The disambiguation strategy based on identification~\cite{m3pl,CroSel} treats the true labels of samples as latent variables and iteratively optimizes the objective function of these latent variables. Based on self-training disambiguation strategies~\cite{parse,mlpll,DongH0Z23,liulearning}, pseudo-labels are used as the model's supervisory information for training, e.g., PRODEN~\cite{proden} re-normalizes the classifier's output and PiCO~\cite{pico} introduces contrastive learning.
Despite the promise, existing PLL work mostly assumes a static data distribution, which is typically not available in practice. 
Therefore, we introduce incremental learning to investigate IPLL that is more suitable for real-world scenarios.

\section{Background}

\subsection{Problem Definition} 
The goal of IPLL is to sequentially learn a unified model from ambiguous supervised datasets and classify unseen test samples of all classes that have been learned so far. Formally, assume we are given a stream of datasets $\{{\cal D}_t\}_{t=1}^T$, where each subset ${\cal D}_t = \{ {({\bm{x}_i^t},{{\cal S}_i^t})}\} _{i = 1}^{{N_t}}$ contains $N_t$ samples. 
Here, ${\bm{x}^t_i} \in {\cal X}$ represents a sample in the input space $\mathbb{R}{^d}$. 
Different from the supervised setup where the ground truth $y_i$ is known, we follow the setup of previous PLL studies~\cite{proden2,proden,pico} and allow the annotator to assign a rough candidate label set $\mathcal{S}_i^t\subset\mathcal{Y}_t$ containing the true label, i.e., $y_i^t\in\mathcal{S}_i^t$. 
For the label space, we consider a \textit{blurry} incremental learning \cite{blurry} that can be ubiquitous in real-world applications\footnote{The classic IL setup simply assumes the label spaces have no intersection, which is less practical and the label ambiguity issue is typically not significant. }, whose data distribution demonstrates:
(i)-a majority of samples with new labels emerge incrementally $\mathcal{Y}_t=\mathcal{Y}_{t-1}\cup\mathcal{Y}^\text{new}_t$;
(ii)-each subset ${{\cal D}_t}$ potentially contains data samples from all labels in $\mathcal{Y}_t$.
At step $t$, the goal of IPLL is to train a model $f$ from the new dataset $\mathcal{D}_t$ without interfering with previous data, where $f$ consists of the feature extractor backbone $\phi$ with a fully-connected layer upon it. During training, since the ground-truth label is not accessible, we assign each sample $\bm{x}_i$ a pseudo-label vector $\bm{p}_i$ and update the model by cross-entropy loss:
\begin{equation}
{\mathcal{L}_{\text{ce}} = -\frac{1}{N_t} \sum_{i=1}^{N_t} \sum_{j=1}^{{|\mathcal{Y}}_t|} p_{ij} \log(f_j(\bm x_i))}
\end{equation}

In the remainder of this work, we omit the task index $t$ when the context is clear.

\subsection{Prototype Generation} 
Recall the crucial challenge of IPLL is to balance the perception of new and old classes. 
We observe that while individual samples exhibit strong ambiguity, prototypes derived from sample aggregation can serve as stable and non-parameterized proxies to guide the model in \textit{identifying new patterns} and \textit{memorizing old tasks}. 
Formally, at the end of the $t$-th training task, we generate prototypes for class $c\in\mathcal{Y}_t$ by feature averaging ${\bm{\mu} _c} = {\frac{1}{|{\bm P_c}|}\sum {\bm P_c} }$, where ${\bm P_c} = \{ \phi ({\bm{x}_i})|c = \mathop {\arg \max }_{j \in {{\cal S}_i^t}} {f_j}({\bm{x}_i})\}$ represents the feature set of samples whose classifier prediction is class $c$. 
In later rounds, holding the belief that the model always produces accurate predictions on old classes, we momentum update the prototypes during the training procedure:
\begin{equation}
\label{eq:propt}
{\bm{\mu} _c} = {\gamma {\bm{\mu} _c} + (1 - \gamma )\frac{1}{{|{\bm P_c}|}}\sum {{\bm P_c}} }, 
\end{equation}
where $\gamma>0$ is a hyperparameter. In what follows, we elaborate on how prototypes help improve both disambiguation and memorization ability.

\begin{figure*}
  \centering
  \includegraphics[width=0.8\linewidth]{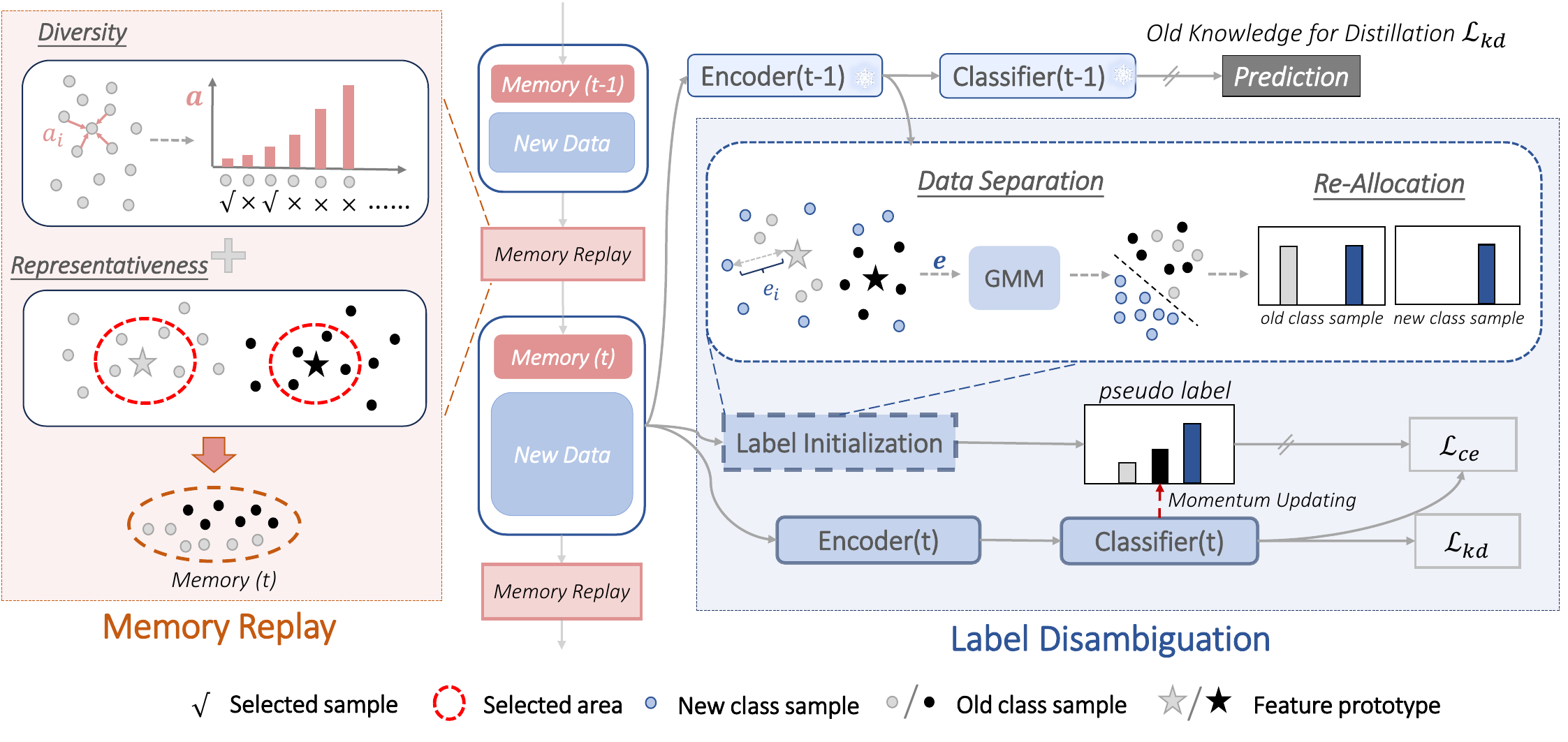}
   \caption{Overall framework of PGDR. The label disambiguation module divides new and old class samples based on prototypes and assigns different labels. PGDR then combines the momentum-updated pseudo-labels to achieve label disambiguation. After completing the task, the memory replay module is utilized to filter samples for subsequent training to mitigate forgetting.}
   \label{fig:fig2}
\end{figure*}

\section{Proposed Method}
\label{sec:method}

In this section, we describe our novel \textbf{P}rototype-\textbf{G}uided \textbf{D}isambiguation and \textbf{R}eplay algorithm (PGDR) in detail. Overall, it consists of two components: (i)-a \textit{prototype-guided label disambiguation} module that first performs old/new sample separation according to their distance to the prototypes, allowing the subsequent task-aware pseudo-labeling; (ii)-a \textit{memory replay} module that constructs a sample pool containing both diverse and representative samples. 
The overall training scheme of PGDR is outlined in Figure~\ref{fig:fig2}.

\subsection{Prototype-Guided Label Disambiguation}
\label{sec:sec4.1}

To handle the label ambiguity, the most seminal PLL algorithms \cite{proden} adopt a self-training paradigm that elicits pseudo-labels from the model outputs. However, in the IPLL setup, such a strategy can be problematic since the classifier can be largely biased after the previous rounds of training. Thus, the disambiguation process would be disrupted since even new class samples can receive very confident predictions on old classes. 

\paragraph{Old/New Data Separation. }
To address the bias, we introduce a prerequisite step that separates samples into distinct subsets for old and new classes. 
Specifically, we first construct a distance measure set of those samples containing at least one old class,
\begin{equation}
\label{eq:la1}
\mathcal{A}=\{e_i=\min_{j\in {\cal S}_i^t \cap {\cal Y}_{t-1}} ||\phi(\bm{x}_i)-\bm{\mu}_j||_2, \text{ if } {\cal S}_i^t \cap {\cal Y}_{t-1}\neq\varnothing\}.
\end{equation}

Our intuition is that, the prototypes fully condense the knowledge of old classes, and thus, samples from old classes are located closer to their prototypes than new ones. 
To achieve fully automated separation, we draw inspiration from the noisy label learning literature \cite{DivideMix} to fit a two-component Gaussian Mixture Model (GMM) on $\mathcal{A}$.
Let $w_i = p(g | e_i)$ represent the probability of $\bm x_i$ belonging to the Gaussian component with smaller mean $g$, which can also be deemed as its old task probability. 
Then, we collect those samples whose $w_i$ exceeds the threshold $\alpha$ as ``\textit{potentially}'' old class data. 
The remaining samples belong to the new class sample set ${\cal D}^{\text{new}}_t$, i.e., ${\cal D}^{\text{new}}_t = {\cal D}_t \setminus {\cal D}^{\text{old}}_t$. 

\paragraph{Candidate Label Re-Allocation. } We re-allocate the candidate label sets according to our data separation procedure,
\begin{equation}
\label{eq:la3}
{{\cal S}_i'} = \begin{cases}
\{c^*_i\} \cup {\cal Y}_t^\text{new} \cap {\cal S}_i^t &\text{if  } \bm x_i\in {\cal D}^{\text{old}}_t, \\
{\cal Y}_t^\text{new} \cap {\cal S}_i^t &\text{otherwise},
\end{cases}
\end{equation}
where $c^*_i$ is a class that corresponds to the nearest prototype.
Our rationale is that: (i)-for a sample is far away from the prototypes (i.e. from ${\cal D}^{\text{new}}_t$), it is highly probable to be a new class sample, and thus, we discard all its old candidate labels; (ii)-for those identified as from old classes, it is either an old class sample locates around one old class prototype (i.e. prototype of $c^*_i$) or actually a new class sample. Hence, we employ the old class prototypes as a disambiguator to remove confusing old classes, but preserve all candidates on new classes.
After this process, the model bias turns into a \textit{good property}---it allows pre-disambiguate old classes. Thus, the model can focus on addressing the ambiguity among new classes. Notably, we conduct the separation procedure once before training for each task to achieve differential guidance for the model.

\paragraph{Momentum-based Pseudo-labeling. } 
After that, we initialize the pseudo-labels $\bm{p}_i$ by a uniform probability on the newly allocated candidate label set:
\begin{equation}
\label{eq:la4}
{ p_{ij}} = \frac{1}{\lvert {\cal S}_i'\rvert} \mathbb{I} (j\in{\cal S}_i').
\end{equation}

In subsequent rounds, we assume the classifier can be increasingly accurate. Thus, we devise a momentum-based mechanism to update the $\bm p_i$: 
\begin{equation}
\label{eq:la5}
{\bm{p}_i} \leftarrow \beta {\bm{p}_i} + (1 - \beta ){\bm{z}_i},
\end{equation}
where $\bm z_i$ is a one-hot vector, i.e., ${z_{ij}} = \mathbb{I}(j=\arg \mathop {\max }_{j \in {{\cal S}_i^t}} {f_j}({\bm{x}_i}))$, and $\beta$ is a hyperparameter.
Practically, we assign a large $\beta$ at the beginning of each task due to the unreliable model prediction of new classes. As the prediction on new classes becomes more accurate, we set a smaller $\beta$ value to facilitate the convergence of the pseudo-labels.



\subsection{Representative and Diverse Memory Replay}
\label{sec:sec4.2}
While our disambiguation module prevents the model from label confusion, the current task is still dominated by new class samples. As the pseudo-labels become increasingly precise, the model tends to be reversely biased towards new patterns and quickly forgets the old knowledge. 
Without stabilizing the old knowledge, the training procedure may be fairly unstable, leading to degraded performance.
To this end, we further introduce a memory replay module to alleviate forgetting, which comprises representative and diverse samples. 

\paragraph{Distance-based Representativeness.}
We assume that most samples can be purified by our disambiguation module at the end of each task. Accordingly, we believe those samples around the prototypes are accurate and can represent well the whole clusters.
Formally, denote the whole memory by $\mathcal{M}^t$ at the $t$-th round, with a maximum limit of $m$ for storing samples. We first concatenate the training set with the previous memory by $\mathcal{D}_t'=\mathcal{D}_t\cup\mathcal{M}^{t-1}$. Then, we refer to prior research \cite{icarl} and perform per-class distance-based selection that collects samples having the shortest distance to their prototype:
\begin{equation}
\label{eq:la7}
\begin{split}
{{\cal M}^t_r} = \cup_{c\in\mathcal{Y}_t} \{{\bm x_i}|&c=\arg \mathop {\max }_{j \in {{\cal S}_i^t}} {f_j}({\bm x_i}),\\
&\text{ rank}_{}(d_i^c) \leq {N_r} \text{, and } {\bm x_i}\in\mathcal{D}_t'\},
\end{split}
\end{equation}
where $d_i^c = ||\phi ({\bm x_i}) - {\bm \mu _{c}}||_2$. That is, we split the whole set according to the predicted categories and then select $N_r$ most representative samples to the episodic memory. 

\paragraph{Neighborhood-based Diversity.} 
Apart from those representative samples, we further collect a few samples that can better preserve the diversity of old class distribution. 
In our IPLL setup, this should be done more carefully due to the label ambiguity.
We propose selecting a sample with two characteristics. First, it should lie in a smooth local manifold to avoid including falsely disambiguated samples. 
Second, it does not lie in a local region near those already selected samples and carries mutually different patterns. 
Formally speaking, we calculate the sum of distance measure from a sample to its $K$-nearest neighbor:
\begin{equation}
\label{eq:la8}
\begin{split}
{a_i} = \sum_{j \in \mathcal{N}_K({\bm{x}_i})} {{d_{ij}}},
\end{split}
\end{equation}
where $\mathcal{N}_K({\bm{x}_i}) = \{ {\bm{x}_j}|\text{rank}({d_{ij}}) \leq {K},j \ne i\}$.
In other words, we believe those samples with lower $a_i$, which indicates they have many close neighbors, are typically located in a relatively smooth region.

Next, we select a sample with the smallest $a_i$ while being not a part of the already chosen sample's neighbors:
\begin{equation}
\label{eq:la10}
\mathcal{M}_k^t = \mathcal{M}_k^t\cup\{\bm{x}_i|a_i=\min_{\bm{x}'\in \mathcal{D}_t'}a'\text{,  and }  \bm{x}_i\notin\cup_{\bm{x}'\in \mathcal{M}^t_k}\mathcal{N}_K(\bm{x}')\}.
\end{equation}

In our implementation, for each class, we first select up to ${N_d}$ diverse samples and then, select $N_r=m/|{{\cal Y}_t}| - {N_d}$ representative samples, resulting in ${\cal M}^t={\cal M}^t_r \cup {\cal M}^t_k$. This provides richer class information for subsequent learning, mitigating the forgetting of old classes. ${\mathcal{L}_\text{ce}}$ also includes this portion of samples

\paragraph{Knowledge Distillation Regularization.} 
In the training phase of the new stage, we utilize replay data ${\cal M}^{t-1}$ and the current data ${\cal D}_t$ for training. 
Following \cite{icarl,wa}, we employ a knowledge distillation loss to alleviate forgetting,
\begin{equation}
\label{eq:la11}
{\mathcal{L}_\text{kd}} = -\frac{1}{|\mathcal{D}'_t|} \sum_{i=1}^{|\mathcal{D}'_t|} \sum_{j=1}^{{|\mathcal{Y}}_{t-1}|}{f^{old}_j({\bm{x}_i})\log (f_j({\bm{x}_i}))}.
\end{equation}

That is, we regularize the new model to mimic the prediction behavior of the old model, and thus, the old knowledge is obviously preserved.

\subsection{Practical Implementation}

\paragraph{Robust Training with Self-Supervised Learning.} In order to further enhance classification performance, we introduce self-supervised learning on the foundation of the disambiguation module and memory replay module. Specifically, we introduce consistency regularization $\mathcal{L}_\text{cr}$~\cite{cr1} to encourage smoother decision boundaries (details in Appendix). 
Finally, the total loss is defined as,
\begin{equation}
\label{eq:la13}
{\mathcal{L}_\text{total}} = {\mathcal{L}_\text{ce}} + {\mathcal{L}_\text{kd}} + {\mathcal{L}_\text{cr}}.
\end{equation}

\paragraph{Bias Elimination for the Testing Phase.} 
In each task, despite having operations to mitigate forgetting, the influence of new class samples on model updates is more profound due to the significantly larger number of new class samples compared to old class samples. Therefore, the classifier exhibits a certain bias. 
In contrast, the feature prototypes retain rich old-class knowledge and are not substantially updated as training progresses.
Consequently, we employ the feature prototypes for sample classification during testing,
\begin{equation}
\label{eq:la14}
{y^*_i} \leftarrow \arg \mathop {\min }\limits_{j \in {{\cal Y}_t}} \left\| {\phi ({\bm{x}_i}) - {{\bm{\mu} }_j}} \right\|.
\end{equation}

As demonstrated empirically, the feature prototype classifier does indeed outperform linear classification during the testing phase; see Appendix.

\begin{table*}[t!]\centering\small
\linespread{1}\selectfont
\setlength{\tabcolsep}{3mm}{
\begin{tabular}{l|cccc|cccc}
\toprule
\makebox[1.17cm][c]{\multirow{3}{*}{\textbf{Method}}} & \multicolumn{4}{c|}{\textbf{CIFAR100}} & \multicolumn{4}{c}{\textbf{Tiny-ImageNet}} \\
\cmidrule(lr){2-5}\cmidrule(lr){6-9}
& \multicolumn{2}{c}{$q=0.1$} & \multicolumn{2}{c|}{$q=0.2$} & \multicolumn{2}{c}{$q=0.1$} & \multicolumn{2}{c}{$q=0.2$} \\
\cmidrule(lr){2-3}\cmidrule(lr){4-5}\cmidrule(lr){6-7}\cmidrule(lr){8-9}
& 10-blurry & 30-blurry & 10-blurry & 30-blurry & 10-blurry & 30-blurry & 10-blurry & 30-blurry \\
\midrule
 PiCO  & 28.00	&28.90 &18.66	&19.34 & 22.62  &19.02  & 10.85 & 9.96 \\
 \midrule[0.01pt]
+iCaRL  &61.59  &62.75  &59.40  &55.25  &41.81  &43.87  &35.27  & 13.98\\
+BiC   &64.11	&64.78  &53.17	&52.93  &33.43  & 34.71  &25.34  &23.41\\
+WA    & \underline{67.31}	& \underline{66.86}  & \underline{64.12}	&59.93  & \underline{44.15}  & \underline{44.33}  &33.80	&31.37\\
+ER-ACE   &62.98  &65.32  &61.32  & \underline{61.29}  &43.14  &43.01  & \underline{40.83}  & \underline{40.79}\\
+ANCL  &65.66	&66.10  &56.04	&48.81  &35.11  &30.31  &23.53	&22.59\\
\cellcolor{gray!30}+PGDR  & \cellcolor{gray!30}\textbf{69.84}	& \cellcolor{gray!30}\textbf{71.68}	& \cellcolor{gray!30}\textbf{68.49}	& \cellcolor{gray!30}\textbf{70.85} &\cellcolor{gray!30}\textbf{46.47}	&\cellcolor{gray!30}\textbf{49.33}	&\cellcolor{gray!30}\textbf{42.03}	&\cellcolor{gray!30}\textbf{41.79}\\
 \midrule[0.01pt]
 PaPi  &24.17	&31.16	&21.59	&22.47 & 20.45	&20.17	&16.34	&10.84\\
 \midrule[0.01pt]
+iCaRL  &60.11  &62.52	&57.69	&58.08  & 38.89	& \underline{37.46}	&22.42	& 20.46\\
+BiC   &61.50	&63.08	&58.40	&56.69  & 41.09	& 35.63	& \underline{25.67}	& \underline{22.82}\\
+WA   &62.75	&\underline{63.59}	&59.73	&\underline{58.10}   & 39.34	&35.82	&20.91	&18.60 \\
+ER-ACE   & 60.76	&62.13	&50.43	&47.17  & 23.60	&21.45	&14.21	&14.10\\
+ANCL  &\underline{63.19}	&63.47	&\underline{60.89}	&57.85  & \underline{42.41}	&35.33	&22.15	&19.26\\
\cellcolor{gray!30}+PGDR  & \cellcolor{gray!30}\textbf{69.09}	& \cellcolor{gray!30}\textbf{69.64} & \cellcolor{gray!30}\textbf{68.47}  & \cellcolor{gray!30}\textbf{66.84} & \cellcolor{gray!30}\textbf{47.85}	& \cellcolor{gray!30}\textbf{49.06}	& \cellcolor{gray!30}\textbf{44.49}	& \cellcolor{gray!30}\textbf{44.64}\\
\bottomrule
\end{tabular}}
\caption{Accuracy comparisons on CIFAR100 and Tiny-ImageNet. The best results are marked in bold and the second-best marked in underline. $q$ represents the degree of label ambiguity. IL methods are equipped with the PLL method PiCO and PaPi.}\label{tab:tab1}
\end{table*}

\begin{table*}[h!]\centering\small
\linespread{1}\selectfont
\setlength{\tabcolsep}{3mm}{
\begin{tabular}{l|cccc|cccc}
\toprule
\makebox[1.17cm][c]{\multirow{3}{*}{\textbf{Method}}} & \multicolumn{4}{c|}{\textbf{CIFAR100-H}} & \multicolumn{4}{c}{\textbf{CUB200}} \\
\cmidrule(lr){2-5}\cmidrule(lr){6-9}
& \multicolumn{2}{c}{$q=0.1$} & \multicolumn{2}{c|}{$q=0.2$} & \multicolumn{2}{c}{$q=0.1$} & \multicolumn{2}{c}{$q=0.2$} \\
\cmidrule(lr){2-3}\cmidrule(lr){4-5}\cmidrule(lr){6-7}\cmidrule(lr){8-9}
& 10-blurry & 30-blurry & 10-blurry & 30-blurry & 10-blurry & 30-blurry & 10-blurry & 30-blurry \\
\midrule
PiCO  &22.85	&26.50	&19.41	&18.16   &21.12	&22.12	&15.25	&13.08\\
\midrule[0.01pt]
+iCaRL &51.69	&47.28	&\underline{51.97}	&42.51  &\underline{46.16}	&41.14	&28.58	&22.37\\
+BiC  &52.75	&53.74	&36.65	&37.32  &43.67	&41.28	&28.11	&22.49 \\
+WA  &53.25  &54.50	&49.26	&46.53  &44.84  &\underline{43.02}	&30.40	&24.14\\
+ER-ACE  &54.22	&54.99	&51.75	&\underline{52.00}  &39.09	&39.48	&23.83	&23.11\\
+ANCL  &\underline{55.59}	&\underline{55.73}	&38.73	&37.37  &44.70	&42.18	&\underline{31.16}	&\underline{24.86}\\
\cellcolor{gray!30}+PGDR &\cellcolor{gray!30}\textbf{60.35}	&\cellcolor{gray!30}\textbf{59.78}	&\cellcolor{gray!30}\textbf{56.85}	&\cellcolor{gray!30}\textbf{54.93} &\cellcolor{gray!30}\textbf{48.83}	&\cellcolor{gray!30}\textbf{49.26}	&\cellcolor{gray!30}\textbf{36.11}	&\cellcolor{gray!30}\textbf{34.04}   \\
\midrule[0.01pt]
PaPi & 18.25	&26.06	&15.87	&17.13	&20.82	&21.87	&16.40 	&14.09\\
\midrule[0.01pt]
+iCaRL  & 47.06	&48.71	&38.37	&\underline{38.93}	&44.61	&38.81	&29.54	&23.62\\
+BiC  & 49.14	&49.01	&41.88	&38.79	&44.70 	&40.60 	&33.04	&\underline{26.47}  \\
+WA & 49.15	&47.03	&38.82	&37.41	&\underline{45.57}	&41.78	&32.29	&25.74\\
+ER-ACE   & 47.27	&44.19	&28.41	&24.80	&43.23	
&37.59	&25.97	&22.22\\
+ANCL  & \underline{51.73}	&\underline{51.09}	&\underline{42.56}	&37.89	&45.27	&\underline{42.56}	&\underline{33.50} 	&25.45 \\
\cellcolor{gray!30}+PGDR &\cellcolor{gray!30}\textbf{58.70}	&\cellcolor{gray!30}\textbf{58.50}	&\cellcolor{gray!30}\textbf{56.75}	&\cellcolor{gray!30}\textbf{55.46} &\cellcolor{gray!30}\textbf{46.30}	&\cellcolor{gray!30}\textbf{47.06}	&\cellcolor{gray!30}\textbf{33.72}	&\cellcolor{gray!30}\textbf{27.19}
\\
\bottomrule
\end{tabular}}
\caption{Accuracy comparisons on CIFAR100-H and CUB200. The best results are marked in bold and the second-best marked in underline. $q$ represents the degree of label ambiguity. IL methods are equipped with the PLL method PiCO and PaPi.}\label{tab:tab2}
\vspace{-0.1cm}
\end{table*}

\section{Experiments}
\label{sec:experiments}

\subsection{Experimental Settings}
\paragraph{Datasets.} We perform experiments on CIFAR100 \cite{cifar100} and Tiny-ImageNet~\cite{tinyimagenet}. Additionally, we further conduct experiments on CUB200~\cite{cub200}. For the experimental setup of IPLL, we reference the settings of PLL~\cite{proden} and IL~\cite{lwf,wa}. We generate partially labeled datasets by manually flipping negative labels $\bar y \ne y$ to false-positive labels with probability $q = P(\bar y\in {\cal Y}_t| \bar y\ne y)$. In the $t$-th task, all $|{\cal Y}_t|-1$ negative labels have a uniform probability to be false positive and we aggregate the flipped ones with the ground-truth to form the candidate set ${\cal S}_i^t$. The flipping probability $q$ is set at 0.1 and 0.2. 
We randomly partition all classes into 10 tasks, i.e., $T=10$. In the first task, there are exclusively new classes, while in the remaining $T-1$ tasks, a substantial number of samples belong to new classes, with relatively fewer samples belonging to old classes. To be specific, for a certain category of samples, $W\%$ of the samples emerge as new class samples in the $t$-th task, while the remaining $(100-W)\%$ of the samples uniformly appear as old class samples in each subsequent task $\{t+1, t+2, ..., T\}$. We consider the degree of blending new and old class data to be ``(100-$W$)-blurry'' 
and set the value of $W$ to 90 (10-blurry) and 70 (30-blurry).

\paragraph{Baselines.} We compare PGDR with two PLL methods: 
 \textbf{PiCO}~\cite{pico} 
and \textbf{PaPi}~\cite{papi}. We also discuss \textbf{PRODEN}~\cite{proden} in Appendix.
We compare PGDR with five incremental learning methods: 1) \textbf{iCaRL}~\cite{icarl}; 2) \textbf{BiC}~\cite{bic}; 3) \textbf{WA}~\cite{wa}; 4) \textbf{ER-ACE}~\cite{erace}; 5) \textbf{ANCL}~\cite{ancl}.
To ensure experimental fairness and credibility, we combine these five incremental learning methods with PiCO and PaPi to achieve basic label disambiguation.
To objectively evaluate our method, we respectively plug the contrastive learning module of PiCO and the Kullback-Leibler divergence of PaPi into our method.

\paragraph{Evaluation metrics.} We utilize the average incremental accuracy as the performance evaluation metric~\cite{icarl}. At the $t$-th task, the incremental accuracy represents the classification accuracy ${A_t}$ of the model on the currently seen classes. The average incremental accuracy is denoted as $\bar A = ({1 \mathord{\left/
 {\vphantom {1 T}} \right.
 \kern-\nulldelimiterspace} T})\sum\nolimits_{i = 1}^T {{A_i}}$.

 \paragraph{Implementation Details.} We employ ResNet-18 for feature extraction. The network model is trained using SGD with a momentum of 0.9. Trained for 200 epochs on the CIFAR100, and trained for 300 epochs on the Tiny-ImageNet. The learning rate starts at 0.1 for PiCO and 0.01 for PaPi. Batch sizes are set to 256 and 128 for the CIFAR100 and Tiny-ImageNet, with a maximum sample storage limit of $m$ of 2000. For prototypes, a moving average coefficient $\gamma$ of 0.5 is used. In the sample selection for the storage phase, we set the number of nearest neighbors $K$ to 10, with a maximum limit $N_d$ for diverse sample storage of $0.67*m/|{\cal Y}_t|$. In the label disambiguation stage, the threshold ${\alpha}$ is set to 0.8. We linearly ramp down $\beta$ from 0.8 to 0.6 to ensure the full utilization of differential labels in the early stages of training. This helps mitigate the introduction of noisy information resulting from inaccurate early predictions.  The experiments regarding PiCO and PaPi reference their experimental parameters.

\subsection{Comparative Results}

\begin{table*}\centering\small
\setlength{\tabcolsep}{3mm}{
\begin{tabular}{@{}c|cccc@{}}
\toprule
\textbf{Ablation} & \textbf{Disambiguation} & \textbf{Memory Replay} & \textbf{10-blurry} & \textbf{30-blurry}\\
\midrule
\rowcolor[gray]{.85} PGDR &\checkmark &\checkmark & \textbf {63.83} & \textbf {64.58} \\
PGDR w MP & Model Prediction  &\checkmark  & 57.94 & 59.13\\
PGDR w PP & Prototype Prediction  &\checkmark  & 59.92 & 62.60  \\\hline
PGDR w/o Memory &$\checkmark$ &$\times$ &23.18 & 23.94  \\
PGDR w Random &\checkmark & Random & 61.34  & 62.60 \\
PGDR w Distance &\checkmark & Prototype Distance & 61.69  & 62.52 \\
\bottomrule
\end{tabular}}
\caption{Ablation study of PGDR on CIFAR100 with $q=0.2$ at IPLL 10 tasks.}\label{tab:tab3}
\vspace{-0.1cm}
\end{table*}

\paragraph{Comparison under IPLL.} As shown in Table~\ref{tab:tab1}, our method significantly outperforms baselines in the IPLL. The PLL methods, PiCO and PaPi, are lower than other solutions due to their difficulty in mitigating forgetting. In addition, on the CIFAR100 dataset with 10 tasks and \( q = 0.1 \), our method outperforms the best baseline incorporating PiCO by {\bf{2.53\%}} (10-blurry) and {\bf{4.82\%}} (30-blurry). Additionally, compared to the best baseline using PaPi, our method shows improvements of {\bf{5.90\%}} (10-blurry) and {\bf{6.05\%}} (30-blurry).
Furthermore, in the case of more noisy candidate labels with $q=0.2$, our method achieves a maximum performance improvement of {\bf{9.56\%}}. We also validate the effectiveness of our method on the more challenging Tiny-ImageNet. 
Our method still demonstrates significant advantages, providing evidence of the superiority of our method in addressing disambiguation and mitigating forgetting.

\noindent \textbf{Comparison on fine-grained datasets.} When semantically similar classes are concentrated in a particular stage, disambiguation becomes more challenging. 
To address this, we reference \cite{pico,solar} and evaluate our method on two fine-grained datasets: 1) CIFAR100 dataset with hierarchical labels, i.e., CIFAR100-H. There are five categories under each superclass, and there is no intersection between different superclasses; 2) CUB200~\cite{cub200} dataset with 200 species of birds. For CIFAR100-H, the new classes in each task are derived from two unseen superclasses in IPLL. The flipping probabilities $q$ are 0.1 and 0.2. As shown in Table~\ref{tab:tab2}, all solutions exhibited a noticeable decrease in performance in the more challenging setting.  
On the CUB200 dataset, when $q$ is set to 0.1, the gaps between our method and the best baseline that incorporates PiCO are {\bf{2.67\%}} (10-blurry) and {\bf{6.24\%}} (30-blurry). 
This thoroughly validates the efficacy of our method, even under highly challenging data settings.

\subsection{Ablation Study}
The additional introduction of the loss function in PGDR hinders visualizing the effects of each module, reducing the interpretability of ablation results. Thus, we present ablation results on PGDR without contrastive learning and Kullback-Leibler divergence to demonstrate our method’s effectiveness.
More experiments can be found in Appendix.

\begin{figure}[t]
\centering
\subfigure[All-class average accuracy.]{
\includegraphics[width=0.23\textwidth]{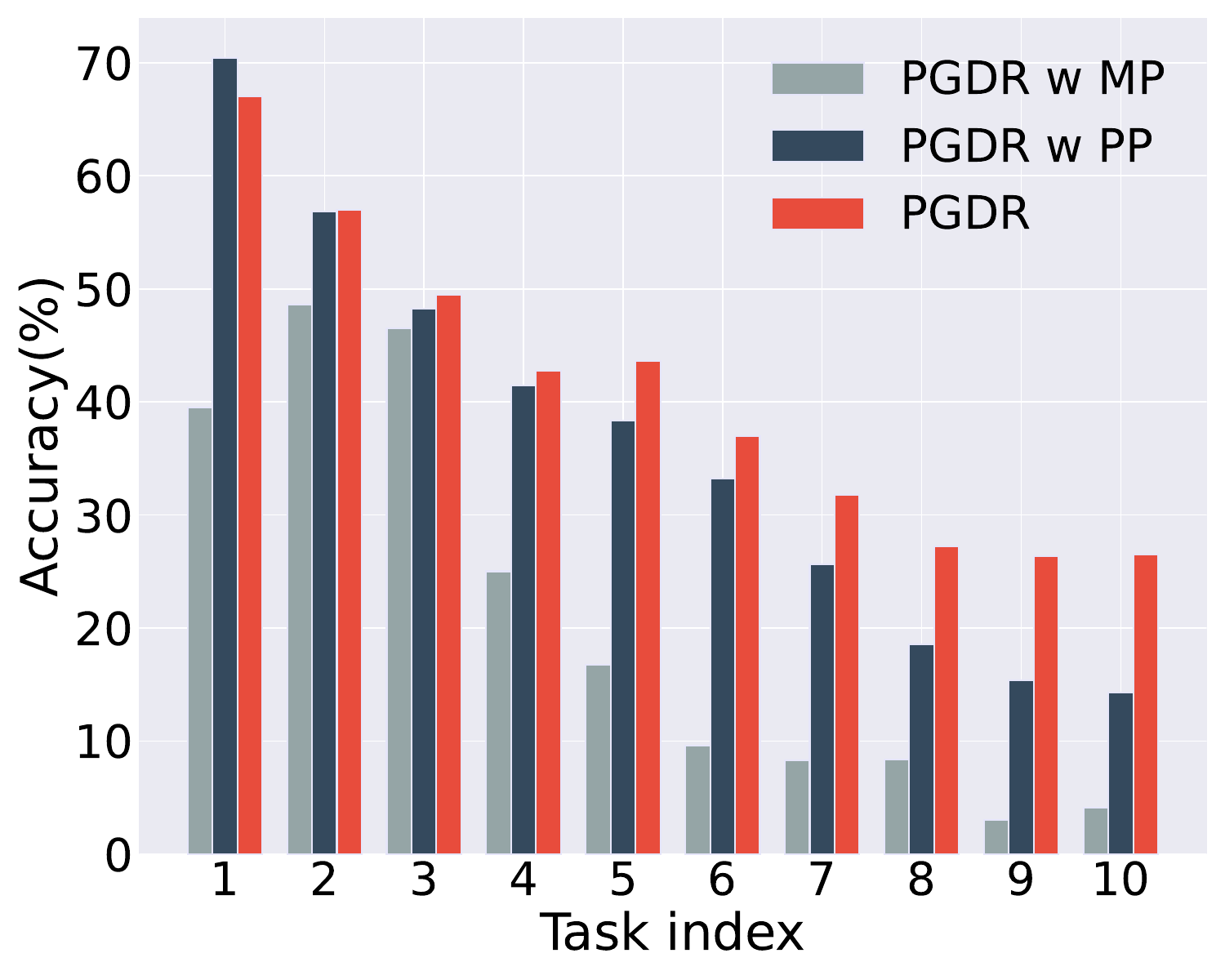}
\label{fig:Disambiguation_new}
}\hspace{-6mm}
\hfill
\subfigure[Old-class average accuracy.]{
\includegraphics[width=0.23\textwidth]{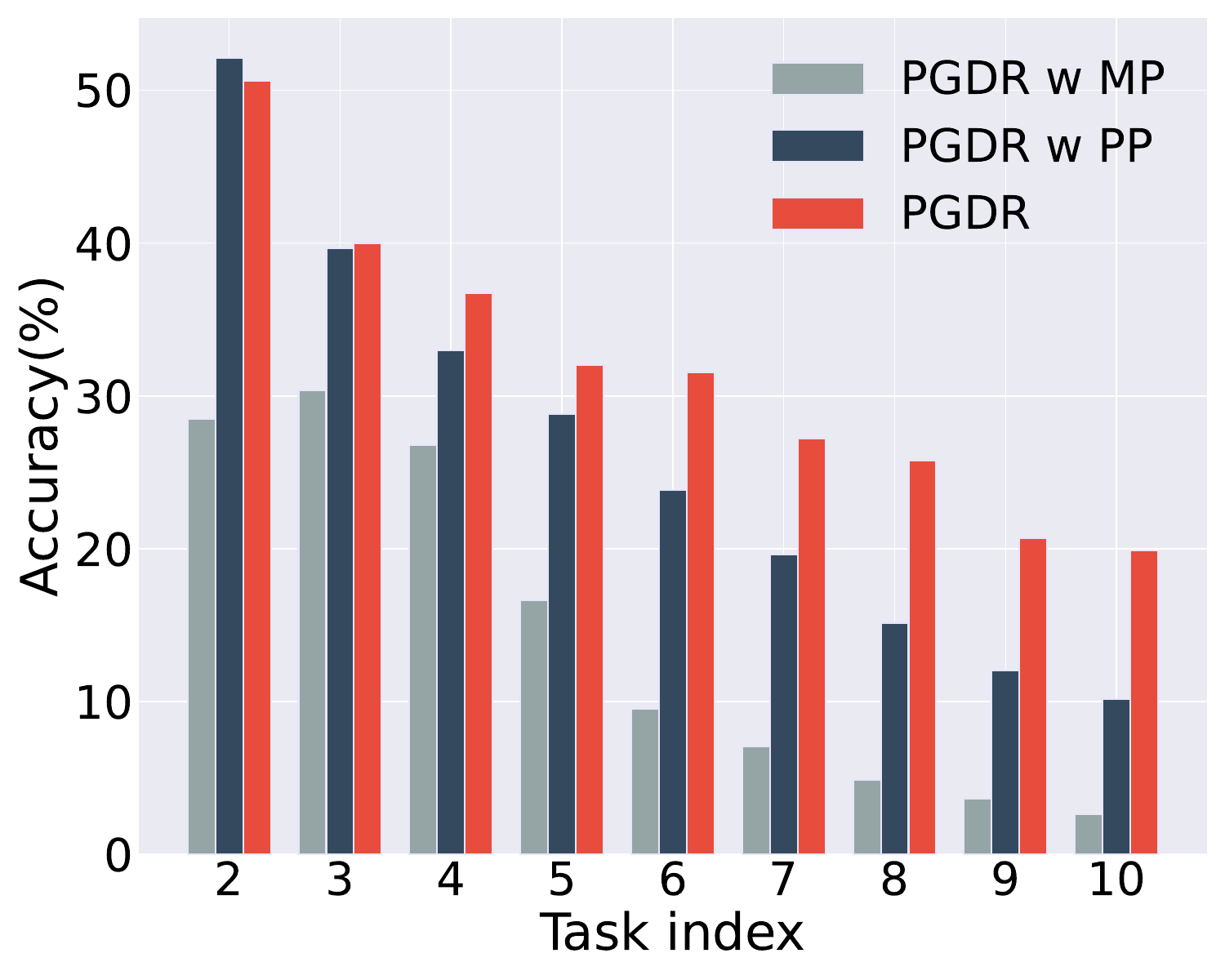}
\label{fig:Disambiguation_old}
}
\caption{Ablation experiment about disambiguation module on Tiny-ImageNet ($q$=0.2)}
\label{fig:fig5}
\end{figure}

\begin{figure}[t]
  \centering
  \subfigure[All-class average accuracy.]{
    \includegraphics[width=0.23\textwidth]{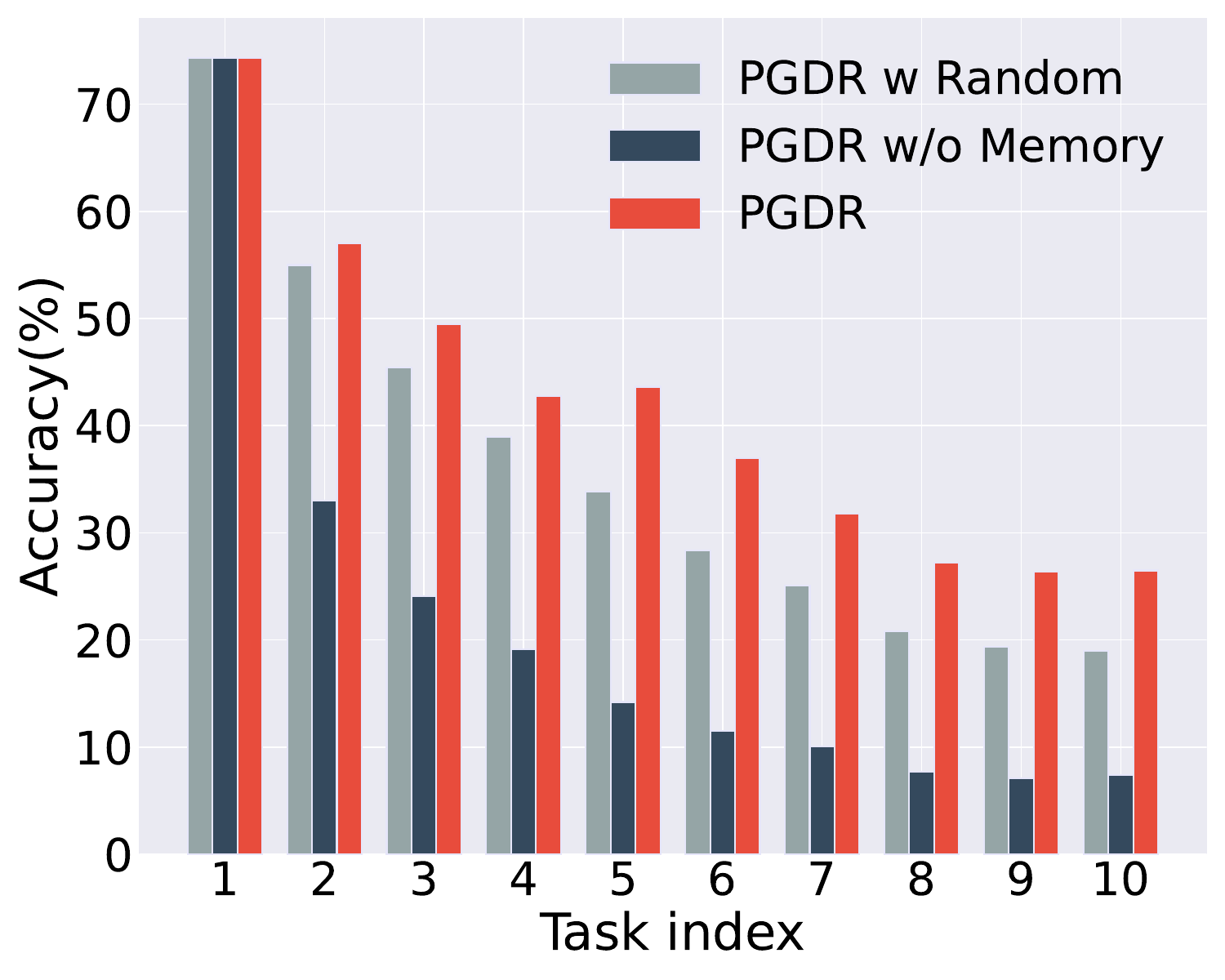}
    \label{fig:mem_all}
  }\hspace{-6mm}
  \hfill
  \subfigure[Old-class average accuracy.]{
    \includegraphics[width=0.23\textwidth]{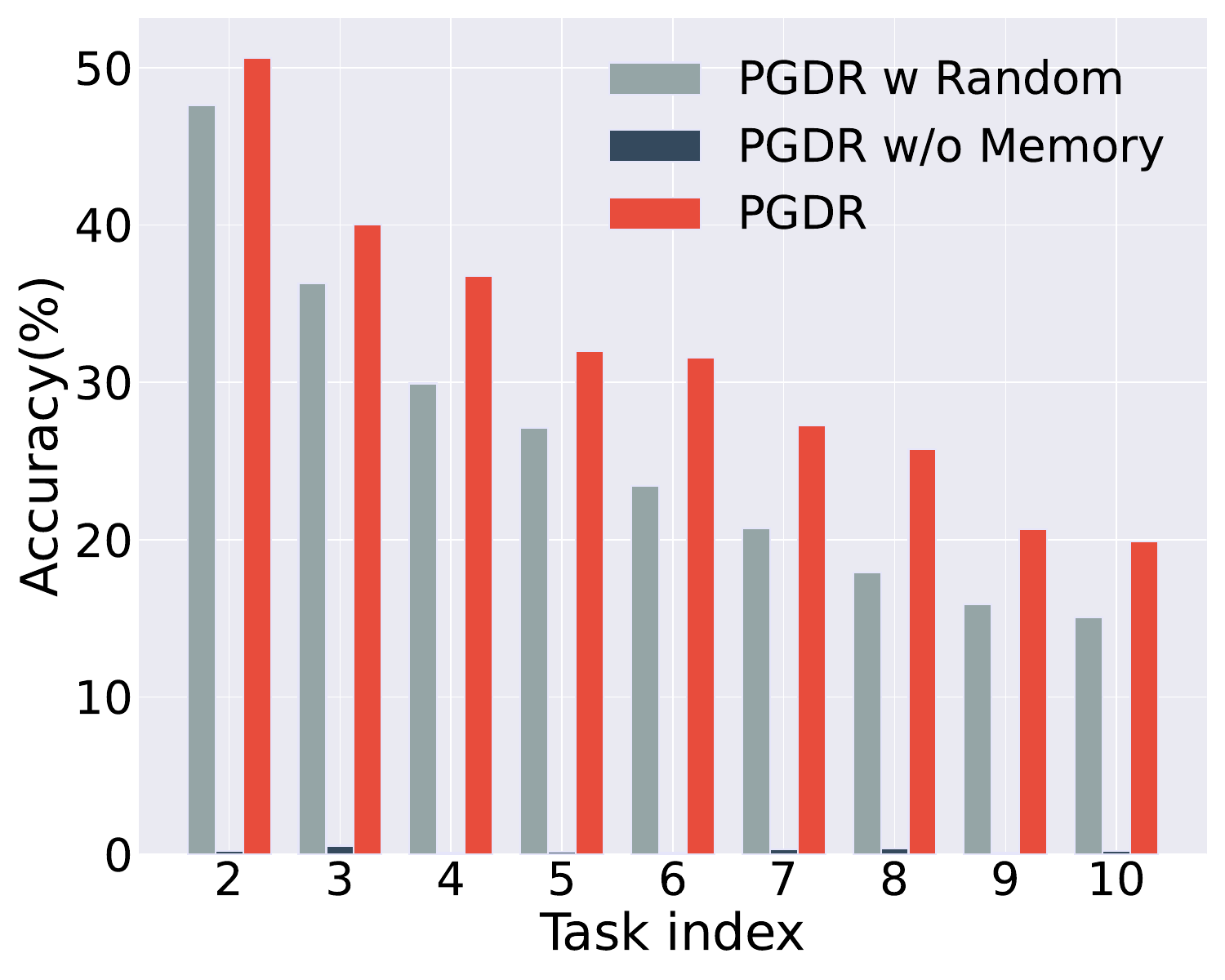}
    \label{fig:mem_old}
  }
  \caption{Ablation experiment about memory replay module on Tiny-ImageNet ($q$=0.2).}
  \label{fig:fig6}
\end{figure}

\paragraph{Effect of Label Disambiguation.} To validate the effectiveness of the label disambiguation module, we choose the variants 1) \textit{PGDR w MP} and 2) \textit{PGDR w PP}; see Appendix. The first variant utilizes the PRODEN label computation method. The second variant utilizes PiCO's labeling method. The Table~\ref{tab:tab3} shows that our disambiguation solution is significantly better than the two variants. To explore further, we visualized category changes under the Tiny-ImageNet 10-blurry (Figure~\ref{fig:fig5}). 
\textit{PGDR w PP} frequent updates to prototypes accumulate model bias, leading to a noticeable decline in classifications when new class data is introduced. In comparison, our solution alleviates knowledge confusion, balancing the model's perception of categories, especially in old classes.

\paragraph{Different Memory Replay Strategies.} We further try different memory update strategies to validate the superiority of our sample memory replay strategy. We compare PGDR with three variants: 1) \textit{PGDR w/o Memory} that does not store samples; 2) \textit{PGDR w Random} randomly selects some samples for storage; 3) \textit{PGDR w Distance} entails selecting only representative samples. As shown in Table~\ref{tab:tab3}, our method outperforms all the variants, which demonstrates the effectiveness of our representative and diverse selection procedure. We detail the comparison results of each task in Figure~\ref{fig:fig6}, where our method outperforms other variants on all-class average accuracy. We can also observe that \textit{PGDR w Random} struggles to overcome forgetting old classes due to the lack of high-quality replay data, and 
\textit{PGDR w/o Memory} suffers from close to zero old-class accuracies with no memory replay.

\section{Conclusion}
We propose and investigate a novel scenario called \textit{incremental partial label learning} (IPLL), where data appears sequentially, and each sample is associated with a set of candidate labels. This presents even more challenging obstacles, involving the intertwining of label disambiguation and catastrophic forgetting. To address this, we propose a prototype-guided disambiguation and replay algorithm (PGDR), which utilizes class prototypes as proxies to balance the model's perception of different categories.
Moreover, we establish the first IPLL benchmark to lay the foundation for future research.
Through extensive experiments, our method consistently outperformed baselines, demonstrating the superiority of PGDR. We hope our work will draw more attention from the community towards a broader view of tackling the IPLL.

\bigskip

\bibliography{aaai25}

\clearpage

\appendix

\section{More Implementation Details}

In this section, we describe more details of the practical implementation of PGDR.

\subsection{Details of Representation Enhancement}
We utilize consistency regularization~\cite{cr1} to enhance robustness to variations in input, which assumes that a classifier should produce similar class probability for a sample and its local augmented copies. Given an image $\bm x_i$, we adopt two different data augmentation modules \cite{NEURIPS2020_d89a66c7,abs-1909-13719} to obtain a weakly-augmented view $\bm x^w_i$ and a strongly-augmented view $\bm x^s_i$. During the training phase, it is imperative to ensure that the predicted outcomes of the strongly-augmented view $\bm x^s_i$ closely approximate the weakly-augmented view, i.e.,
${\mathcal{L}_{\text{cr}} = --\frac{1}{|\mathcal{D}'_t|} \sum_{i=1}^{|\mathcal{D}'_t|} \sum_{j=1}^{{|\mathcal{Y}}_t|} p_{ij} \log(f_j(\bm x^s_i))}$
, which involves a label $\bm p_i$ based on the prediction from the weakly-augmented view.

\subsection{Details on Label Disambiguation Ablation}\label{L_D}
 PRODEN and PiCO are classical label disambiguation methods. Therefore, we compare both approaches with our method to assess the efficacy of PGDR's label disambiguation module. Our label disambiguation module consists of candidate label re-allocation and pseudo-label update. In the ablation experiments, we replace this module with PRODEN's labeling strategy (\textit{PGDR w MP}) and PiCO's labeling strategy (\textit{PGDR w PP}).
 For \textit{PGDR w MP}, the model predictions are directly used as the pseudo-labels. For \textit{PGDR w PP}, it employs prototype predictions to momentum update pseudo-labels. It is important to note that in \textit{PGDR w PP}, the prototypes used for updating pseudo-labels need to be updated at the end of each epoch.

\subsection{Additional Implementation Details}
In experiments related to PiCO and PaPi, the projection head of the contrastive network is a 2-layer MLP that generates 128-dimensional embeddings. The queue size, which stores key embeddings, is set to a fixed value of 4096. Additionally, cosine learning rate scheduling is implemented. On the CUB200 dataset, the learning rate for all methods is set to 0.01, with 20 epochs and a batch size of 128. Besides, the ANCL utilizes an auxiliary network to improve model performance, and we adopt its BiC variant for comparison. 

\section{Additional Experimental Results}
In this section, we report the additional empirical results of our proposed PGDR framework. Because PLL methods using contrastive learning outperform PLL methods using Kullback-Leibler divergence in most cases, we explore the effect of methods using contrastive learning in different situations. All experiments are conducted on a workstation with 8 NVIDIA A6000 GPUs. The licenses of our employed datasets are unknown (non-commercial).

\begin{table}\small\centering
\begin{tabular}{l|cc}
\toprule%
 \textbf{Method} & \textbf{CIFAR100} & \textbf{Tiny-ImageNet} \\
\midrule
 PiCO &24.26	&20.56 \\
 \midrule
+iCaRL  &58.78	&32.69 \\
+BiC  &57.13	&25.48 \\
+WA   &\underline{61.50}	 &33.59 \\
+ER-ACE   &61.35	&\underline{38.23} \\
+ANCL &56.44	&23.77 \\
\cellcolor{gray!30}+PGDR  & \cellcolor{gray!30}\textbf{63.26}	& \cellcolor{gray!30}\textbf{40.41} \\
\bottomrule%
\end{tabular}
\caption{Accuracy comparisons on non-blurry IPLL.}\label{tab:non-blurry}
\end{table}

 \begin{table}\small\centering
\begin{tabular}{l|cc}
\toprule%
 \textbf{Method} & \textbf{10-blurry} & \textbf{30-blurry} \\
\midrule
 PiCO &31.98	&45.05 \\
 \midrule[0.01pt]
+iCaRL  &61.59	&65.42 \\
+BiC  &65.53	&66.60 \\
+WA   &\underline{66.53}	 &\underline{67.47} \\
+ER-ACE   &65.79	&67.16 \\
+ANCL &65.30	&67.37 \\
\cellcolor{gray!30}+PGDR  & \cellcolor{gray!30}\textbf{67.10}	& \cellcolor{gray!30}\textbf{68.13} \\
\bottomrule%
\end{tabular}
\caption{Accuracy comparisons on CIFAR100 with non-uniform generated candidate labels.}\label{tab:fig_non}
\end{table}

\begin{table}[!t]\small\centering
\begin{tabular}{l|cccc}
\toprule
\makebox[1.17cm][c]{\multirow{2}{*}{\textbf{Method}}} & \multicolumn{2}{c}{\textbf{20 tasks}} & \multicolumn{2}{c}{\textbf{25 tasks}} \\
\cmidrule(lr){2-3}\cmidrule(lr){4-5}
& $q=0.1$ & $q=0.2$ & $q=0.1$ & $q=0.2$ \\
\midrule
 PiCO  &17.98  &16.12  &15.17  &14.07\\
 \midrule[0.01pt]
+iCaRL  &59.75  & 56.70  &59.75  &56.65\\
+BiC   &60.25  &58.62  &58.53  &57.24\\
+WA   &\underline{64.12}  &\underline{60.67}  &\underline{62.49}  &\underline{59.34}\\
+ER-ACE   &60.44  &58.30  &59.33  &56.68\\
+ANCL &59.67  &58.79  &59.77  &58.18\\
\cellcolor{gray!30}+PGDR  & \cellcolor{gray!30}\textbf{65.54}	& \cellcolor{gray!30}\textbf{64.09} & \cellcolor{gray!30}\textbf{63.21}  & \cellcolor{gray!30}\textbf{63.82}\\
\bottomrule
\end{tabular}
\caption{Accuracy comparisons on CIFAR100 under IPLL. The best results are marked in bold and the second-best marked in underline. $q$ represents the degree of label ambiguity.  IL methods are equipped with the PLL method PiCO.}\label{tab:tasks}
\end{table}


\subsection{Results on Non-blurry IPLL}
Referring to the traditional CIL \cite{icarl,wa}, we investigate the effectiveness of PGDR in \textit{non-blurry IPLL}. We set that new samples appearing in each task are derived from new classes, i.e., 0-blurry.
As show in Table~\ref{tab:non-blurry}, under the condition of a flipping probability $q$ of 0.2, our solution outperforms baselines equipped with PiCO. On the CIFAR100 and Tiny-ImageNet datasets, PGDR outperforms the best baseline by {\bf{1.76\%}} and {\bf{2.18\%}}. This highlights the effectiveness of our method in handling disambiguation and mitigating forgetting.

\begin{table*}\centering\small
\vspace{-0.2cm}
\setlength{\tabcolsep}{3mm}{
\begin{tabular}{l|cccc|cccc}
\toprule
\multirow{3}{*}{\textbf{Method}} & \multicolumn{4}{c|}{\textbf{CIFAR100}} & \multicolumn{4}{c}{\textbf{Tiny-ImageNet}} \\
\cmidrule(lr){2-5}\cmidrule(lr){6-9}
& \multicolumn{2}{c}{$q=0.1$} & \multicolumn{2}{c|}{$q=0.2$} & \multicolumn{2}{c}{$q=0.1$} & \multicolumn{2}{c}{$q=0.2$} \\
\cmidrule(lr){2-3}\cmidrule(lr){4-5}\cmidrule(lr){6-7}\cmidrule(lr){8-9}
& 10-blurry & 30-blurry & 10-blurry & 30-blurry & 10-blurry & 30-blurry & 10-blurry & 30-blurry \\
\midrule
 PRODEN  & 25.26 & 32.95 & 24.12 & 24.81 &18.13	&17.86  &18.54	&17.62\\
 \midrule
+iCaRL  & 51.61 & 54.02 & 51.55 & 54.26 & 29.28 & 30.25 &29.34	&30.24\\
+BiC   & 52.92	 & 57.24	& 46.83  & 54.90   & \underline{32.78} & 33.49  &33.26	&33.98\\
+WA    & \underline{54.53}	& \underline{59.57}	& \underline{52.55}	& 53.94  & 32.10 & \underline{35.11}  &33.03	&\underline{34.52}   \\
+ER-ACE   & 48.27	& 48.43	& 48.30	& 48.13 & 29.36 & 29.07   &27.99	&27.38 \\
+ANCL   & 53.87  & 58.61	& 49.23	& \underline{55.60} & 31.56 & 32.67  &\underline{34.03}	&34.27 \\
\cellcolor{gray!30} PGDR  & \cellcolor{gray!30}\textbf{64.40}	& \cellcolor{gray!30}\textbf{65.54}	& \cellcolor{gray!30}\textbf{63.83}	& \cellcolor{gray!30}\textbf{64.58} &\cellcolor{gray!30}\textbf{41.60}	&\cellcolor{gray!30}\textbf{42.39}	&\cellcolor{gray!30}\textbf{38.46}	&\cellcolor{gray!30}\textbf{39.83}\\
\bottomrule
\end{tabular}}
\caption{Accuracy comparisons on CIFAR100 and Tiny-ImageNet. The best results are marked in bold and the second-best marked in underline. $q$ represents the degree of label ambiguity. IL methods are equipped with the PLL method PRODEN.}\label{tab:table7}
\end{table*}

\begin{table*}\centering\small
\setlength{\tabcolsep}{3mm}{
\begin{tabular}{l|cccc|cccc}
\toprule
\multirow{3}{*}{\textbf{Method}} & \multicolumn{4}{c|}{\textbf{CIFAR100-H}} & \multicolumn{4}{c}{\textbf{CUB200}} \\
\cmidrule(lr){2-5}\cmidrule(lr){6-9}
& \multicolumn{2}{c}{$q=0.1$} & \multicolumn{2}{c|}{$q=0.2$} & \multicolumn{2}{c}{$q=0.1$} & \multicolumn{2}{c}{$q=0.2$} \\
\cmidrule(lr){2-3}\cmidrule(lr){4-5}\cmidrule(lr){6-7}\cmidrule(lr){8-9}
& 10-blurry & 30-blurry & 10-blurry & 30-blurry & 10-blurry & 30-blurry & 10-blurry & 30-blurry \\
\midrule
PRODEN &21.46	&29.80	&19.60	&21.46  &15.37	&14.45	&11.64	&10.97\\
\midrule
+iCaRL  &44.98	&41.74	&41.78	&44.32  &23.22	&23.82	&18.44	&15.43\\
+BiC   &42.65	&47.73	&43.50	&44.62  &28.67	&23.04	&18.18	&15.47    \\
+WA    &46.02	&47.73	&\underline{45.74}	&\underline{46.86}  &28.17	&24.23	&19.90	&15.48  \\
+ER-ACE    &41.50	&41.11	&40.43	&40.91   &\underline{40.91}	&\underline{41.86}	&\underline{28.74}	&\underline{27.44} \\
+ANCL  &\underline{47.13}	&\underline{48.96}	&43.47	&45.34  &30.45	&24.20	&18.37	&15.45 \\
\cellcolor{gray!30} PGDR   &\cellcolor{gray!30}\textbf{55.96}	&\cellcolor{gray!30}\textbf{56.47}	&\cellcolor{gray!30}\textbf{54.78}	&\cellcolor{gray!30}\textbf{55.72} &\cellcolor{gray!30}\textbf{48.48}	&\cellcolor{gray!30}\textbf{48.97}	&\cellcolor{gray!30}\textbf{42.51}	&\cellcolor{gray!30}\textbf{41.39}  \\
\bottomrule
\end{tabular}}
\caption{Accuracy comparisons on CIFAR100-H and CUB200. The best results are marked in bold and the second-best marked in underline.
q represents the degree of label ambiguity. IL methods are equipped with the PLL method PRODEN.}\label{tab:table8}
\end{table*}

\subsection{Results with Non-Uniform Data Generation}
In practice, some labels may be more analogous to the true label than others, which makes their probability of label flipping $q$ larger than others. It means that the data generation procedure is non-uniform. Referring to~\cite{pico,solar}, we further test PGDR with a non-uniform data generation process on CIFAR100, with the flipping matrix~\ref{matrix}, where each entry denotes the probability of a label being a candidate. In the current task, there is a greater likelihood of confusion between new classes and old classes. Our method still maintains a strong performance advantage. As shown in Table~\ref{tab:fig_non}, PGDR achieves promising results, outperforming baselines in 10-blurry and 30-blurry.
\begin{equation}\label{matrix}
\begin{bmatrix}
    1 & 0 & 0 & 0 & 0 & 0 & 0 & ... & 0 \\
    0.5 & 1 & 0 & 0 & 0 & 0 & 0 & ... & 0 \\
    0.4 & 0.5 & 1 & 0 & 0 & 0 & 0 & ... & 0 \\
    0.3 & 0.4 & 0.5 & 1 & 0 & 0 & 0 & ... & 0 \\
    0.2 & 0.3 & 0.4 & 0.5 & 1 & 0 & 0 & ... & 0 \\
    0.1 & 0.2 & 0.3 & 0.4 & 0.5 & 1 & 0 & ... & 0 \\
    0 & 0.1 & 0.2 & 0.3 & 0.4 & 0.5 & 1 & ... & 0 \\
    \vdots &&&&...&&&& \vdots\\
    0 & ... & 0 & 0.1 & 0.2 & 0.3 & 0.4 & 0.5 & 1
\end{bmatrix}
\end{equation}

\subsection{Results for Different Number of Tasks}
We conduct experiments to discuss the effectiveness of our method as the number of tasks increases. We set the task quantities to be 20 and 25. 
 As shown in Table~\ref{tab:tasks}, under the 10-blurry condition, our method still maintains an advantage. Specifically, in the case of 20 tasks, we outperform the best baseline by {\bf{1.42\%}} and {\bf{3.42\%}}. In the case of 25 tasks, our method outperforms the best baseline by {\bf{4.48\%}}. This validates the effectiveness of our method in long-term learning.


\subsection{Results of Baselines Equipped with PRODEN}
We further enhance our investigation into the category perception balancing ability of PGDR by incorporating IL baselines with an additional PLL methods: PRODEN~\cite{proden}, a classic PLL method.
As illustrated in Table~\ref{tab:table7}, our method achieves a performance enhancement of {\bf{9.87\%}} (10-blurry) and {\bf{5.97\%}} (30-blurry) on the CIFAR100 dataset, surpassing the best baseline in IPLL with \( q=0.1 \). In a more challenging scenario with \( q=0.2 \), our approach demonstrates an improvement of {\bf{11.28\%}} (10-blurry) and {\bf{8.98\%}} (30-blurry) relative to the best baseline. Additionally, as presented in Table~\ref{tab:table8}, our method outperforms the baselines on fine-grained datasets. This once again validates the effectiveness of PGDR in balancing category perception.

\subsection{Comparison with the SOTA IL Method}
To further validate the performance advantages of our method in IPLL, we discuss the SOTA IL method, C-Flat~\cite{c-flat}. We utilize the WA version of C-Flat for our experiments. As shown in Table~\ref{tab:c-flat}, we systematically combine C-Flat with three PLL methods to mitigate forgetting. Our findings indicate that while the combination of C-Flat with PiCO yields the best results, it still performs worse than PGDR. This reinforces the idea that the significant potential of PGDR.

\subsection{Comparison with the SOTA PLL Method}
We integrate IL techniques with the SOTA PLL method, TERIAL~\cite{terial}, to further validate the effectiveness of PGDR. As shown in Table~\ref{tab:terial}, our method demonstrates superior performance. However, the baselines did not perform better than those equipped with other PLL methods. We suspect this is due to TERIAL causing the model to overly focus on new categories while neglecting the retention of old knowledge, leading to accelerated forgetting of prior information. 

\begin{table}[!t]\small\centering
\begin{tabular}{l|cccc}
\toprule
\makebox[1.17cm][c]{\multirow{2}{*}{\textbf{Method}}} & \multicolumn{2}{c}{\textbf{$q=0.1$}} & \multicolumn{2}{c}{\textbf{$q=0.2$}} \\
\cmidrule(lr){2-3}\cmidrule(lr){4-5}
& 10-blurry  & 30-blurry & 10-blurry  & 30-blurry \\
\midrule
 PRODEN & 25.26 & 32.95 & 24.12 & 24.81 \\
 +C-Flat & 62.89	&63.44	&62.16	&63.15\\
\cellcolor{gray!30} PGDR  & \cellcolor{gray!30}\textbf{64.40}	& \cellcolor{gray!30}\textbf{65.54}	& \cellcolor{gray!30}\textbf{63.83}	& \cellcolor{gray!30}\textbf{64.58} \\
 \midrule
 PiCO & 28.00	&28.90 &18.66	&19.34 \\
 +C-Flat & 69.45	&69.92	&67.49	&65.80\\
 \cellcolor{gray!30}+PGDR  & \cellcolor{gray!30}\textbf{69.84}	& \cellcolor{gray!30}\textbf{71.68}	& \cellcolor{gray!30}\textbf{68.49}	& \cellcolor{gray!30}\textbf{70.85} \\
 \midrule
 PaPi &24.17	&31.16	&21.59	&22.47  \\
  +C-Flat &63.74	&64.75	&60.91	&60.73\\
 \cellcolor{gray!30}+PGDR  & \cellcolor{gray!30}\textbf{69.09}	& \cellcolor{gray!30}\textbf{69.64} & \cellcolor{gray!30}\textbf{68.47}  & \cellcolor{gray!30}\textbf{66.84} \\
\bottomrule
\end{tabular}
\caption{Accuracy comparisons on CIFAR100 under IPLL.}\label{tab:c-flat}
\end{table}

\begin{table}[!t]\small\centering
\begin{tabular}{l|cccc}
\toprule
\makebox[1.17cm][c]{\multirow{2}{*}{\textbf{Method}}} & \multicolumn{2}{c}{$q=0.1$} & \multicolumn{2}{c}{$q=0.2$} \\
\cmidrule(lr){2-3}\cmidrule(lr){4-5}
& 10-blurry  & 30-blurry & 10-blurry  & 30-blurry \\
\midrule
 TERIAL &26.11	&33.37	&24.39	&25.20  \\
 \midrule
+iCaRL &48.10 	&49.89	&44.39	&46.21 \\
+BiC  & 50.55	&49.06	&47.25	&48.68   \\
+WA   &48.42	&49.53	&44.43	&46.88 \\
+ER-ACE   &47.79	&49.04	&46.12	&46.22  \\
+ANCL  &\underline{50.64}	&\underline{51.63}	&\underline{51.45}	&\underline{51.46} \\
\cellcolor{gray!30} PGDR  & \cellcolor{gray!30}\textbf{64.40}	& \cellcolor{gray!30}\textbf{65.54}	& \cellcolor{gray!30}\textbf{63.83}	& \cellcolor{gray!30}\textbf{64.58}   \\
\bottomrule
\end{tabular}
\caption{Accuracy comparisons on CIFAR100 under IPLL. The IL methods are combined with TERIAL.}\label{tab:terial}
\end{table}

\begin{table}[!t]\small\centering
\scalebox{0.8}{
\begin{tabular}{c|ccccc}
\toprule%
 \textbf{Parameter} & \bm{$\beta=0.2$} & \bm{$\beta=0.4$} & \bm{$\beta=0.6$} & \bm{$\beta=0.8$} & \bm{$\beta=1.0$}\\
\midrule[0.25pt]
Accuracy   	&64.08	&63.46	&63.64	&62.73	&39.82\\
\bottomrule%
\end{tabular}}
\caption{Performance of PGDR with label-updating parameter $\beta$ on CIFAR100.}\label{tab:beta}
\end{table}

\begin{table}\small\centering
\scalebox{0.8}{
\begin{tabular}{c|ccccc}
\toprule%
 \textbf{Parameter} & \bm{$\alpha=0.2$} & \bm{$\alpha=0.4$} & \bm{$\alpha=0.6$} & \bm{$\alpha=0.8$} & \bm{$\alpha=1.0$}\\
\midrule[0.25pt]
Accuracy   	&62.54	&64.19	&63.78  &63.83	&63.47\\
\bottomrule%
\end{tabular}}
\caption{Performance of PGDR with disambiguation threshold $\alpha$ on CIFAR100.}\label{tab:alpha}
\end{table}

\begin{figure}[t]
  \centering
  \begin{minipage}[t]{0.3\textwidth}
    \raggedright
    \includegraphics[width=1\textwidth]{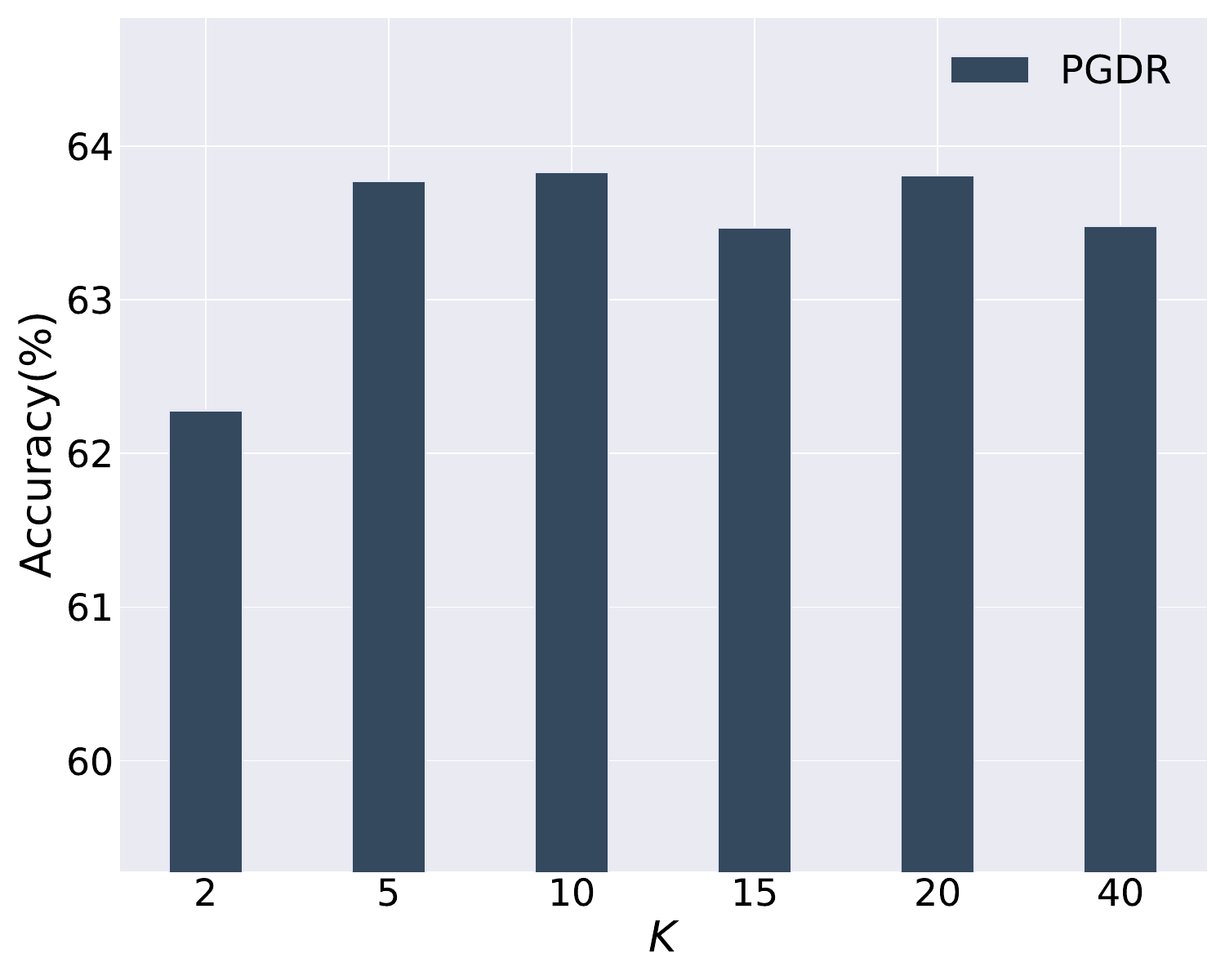}
    \caption{Performance of PGDR with varying $K$ on CIFAR100 (10-blurry).}
    \label{fig:k}
  \end{minipage}
  \begin{minipage}[t]{0.3\textwidth}
  \raggedright
    \includegraphics[width=1\textwidth]{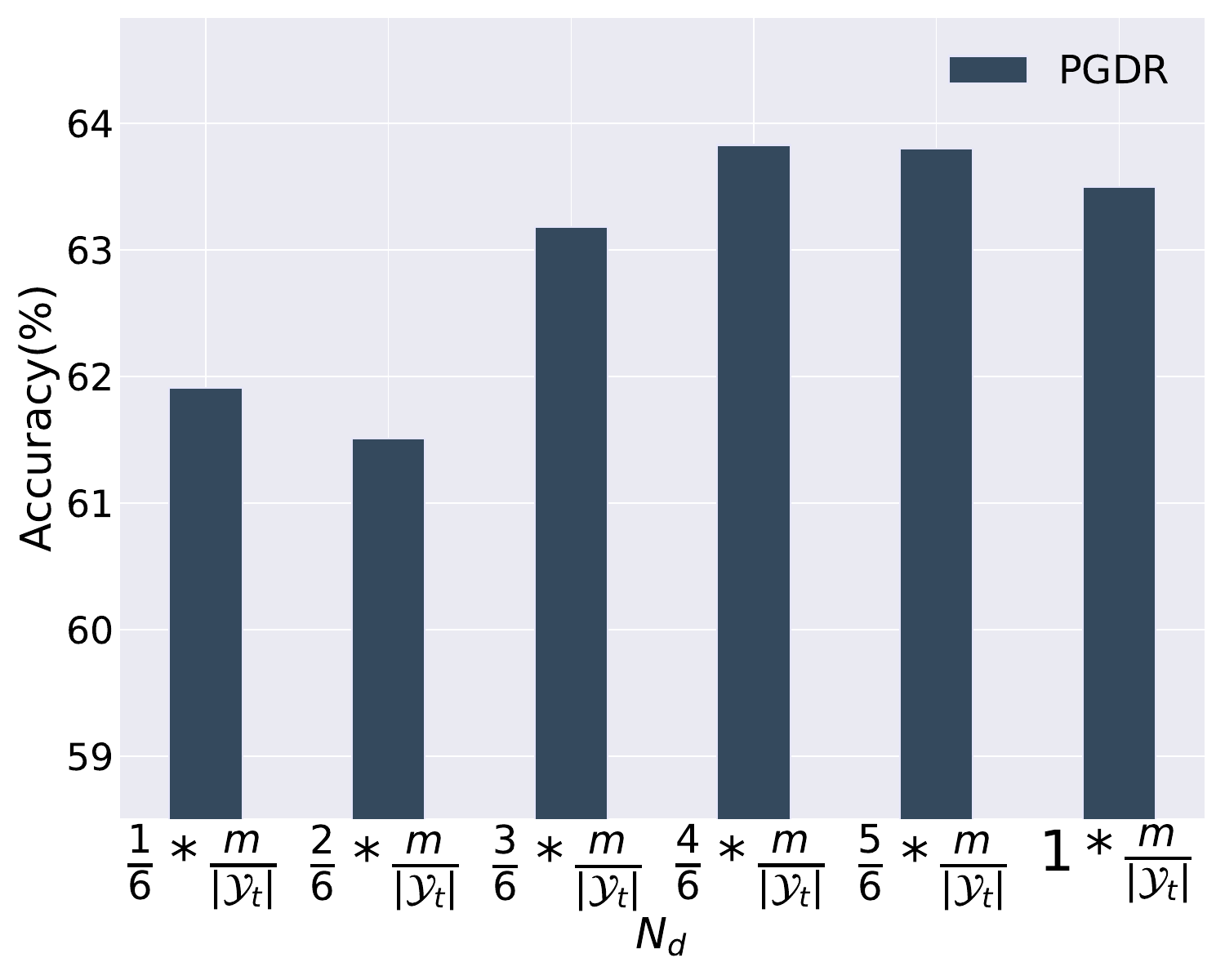}
    \caption{Performance of PGDR with varying $N_d$ on CIFAR100 (10-blurry).}
    \label{fig:n_d}
  \end{minipage}
\end{figure}

\begin{figure*}[t]
  \centering
  \begin{minipage}[t]{0.3\textwidth}
    \raggedright
    \includegraphics[width=1\textwidth]{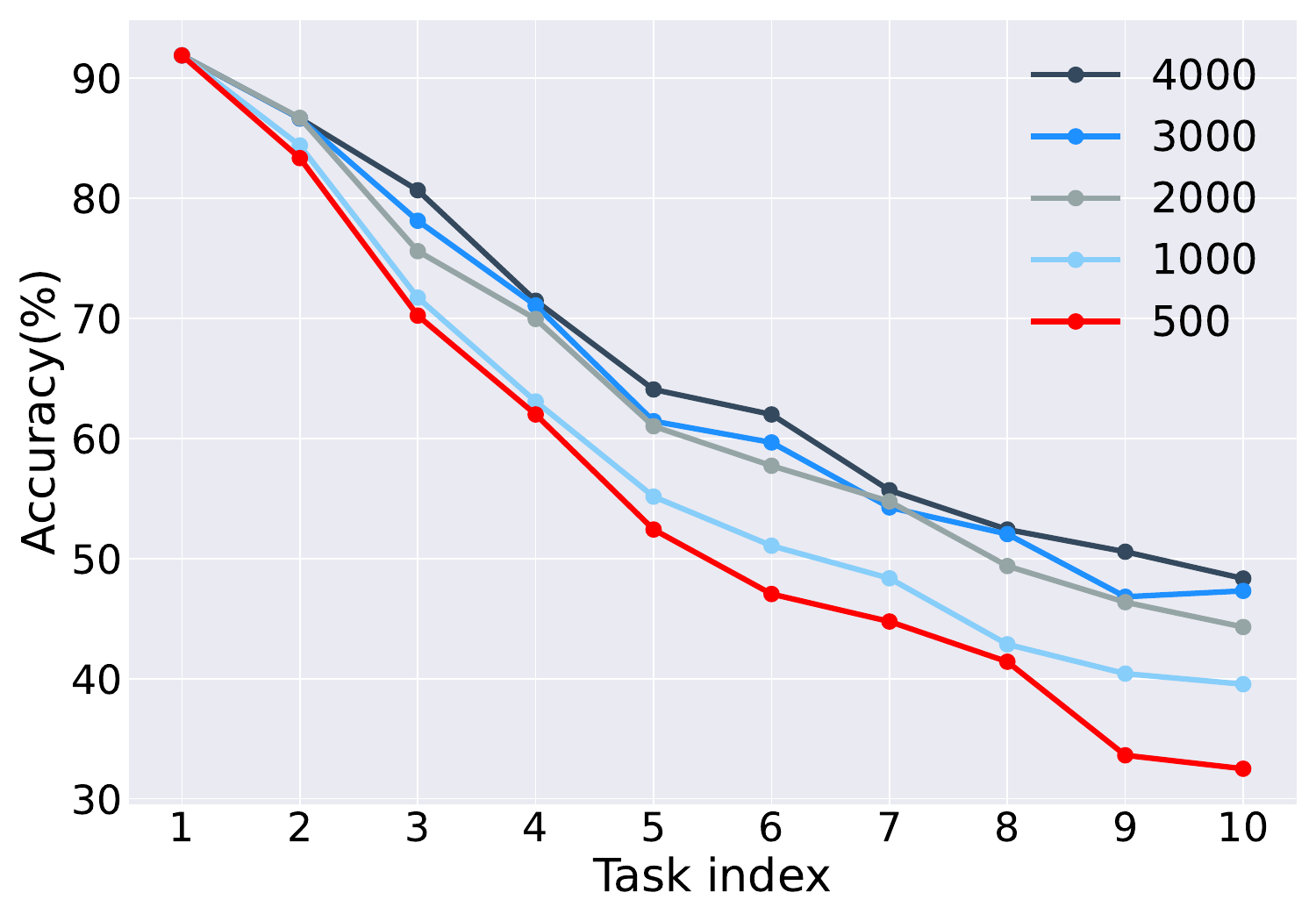}
    \caption{Performance of PGDR with varying $m$ on CIFAR100 (10-blurry).}
    \label{fig:m}
  \end{minipage}
  \hspace{.4in}
  \begin{minipage}[t]{0.3\textwidth}
  \raggedright
    \includegraphics[width=1\textwidth]{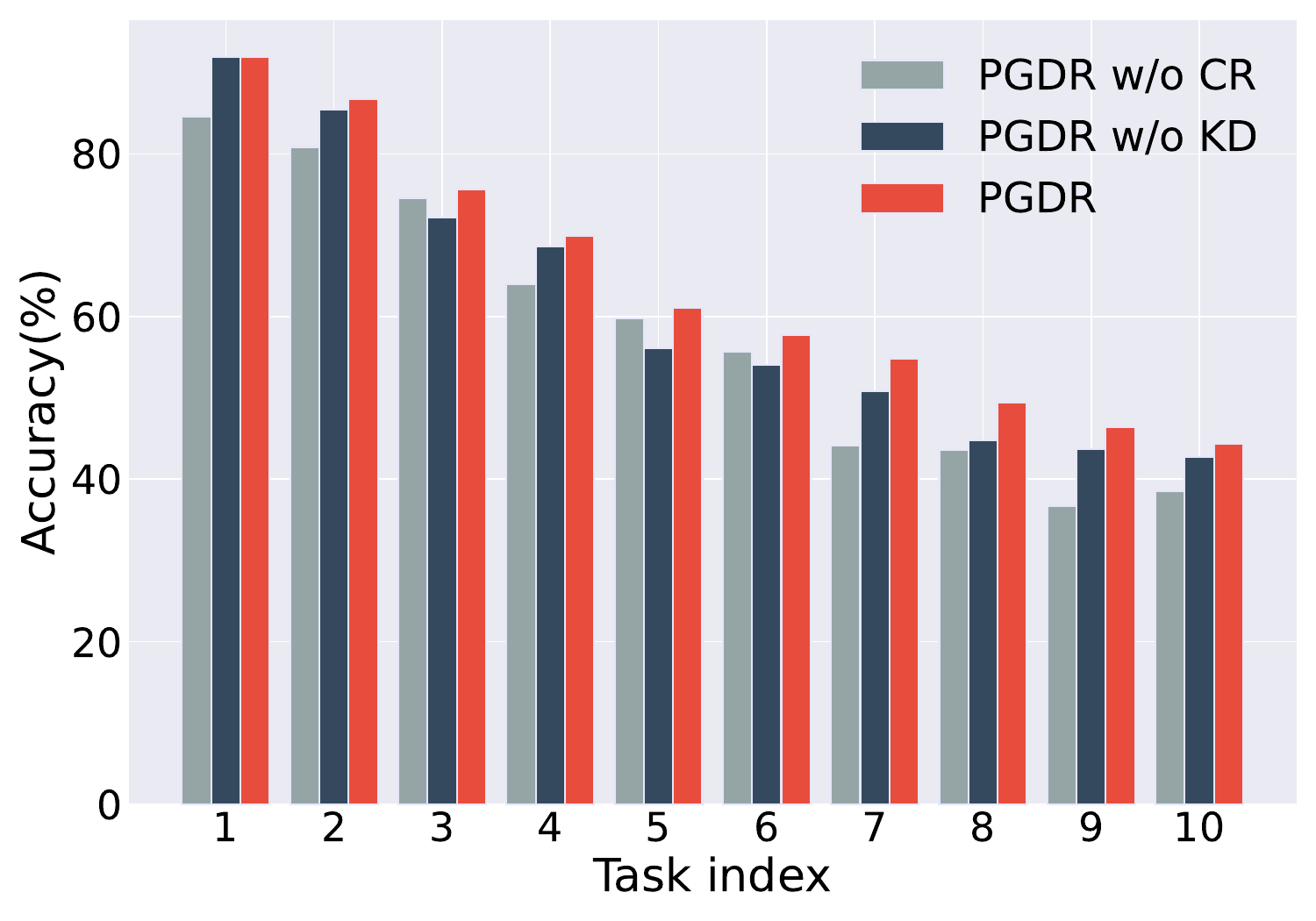}
    \caption{Performance comparison of PGDR, \textit{PGDR w/o CR}, and \textit{PGDR w/o KD} on CIFAR100 (10-blurry).}
    \label{fig:cr_kd}
  \end{minipage}
  \label{fig:m_cr_kd}
\end{figure*}

\begin{figure*}[t!]
  \centering
  \subfigure[All-class average accuracy.]{
    \includegraphics[width=0.32\textwidth]{source/Disambiguation_2.pdf}
    \label{fig:Disambiguation_all}
  }
  \hfill
  \subfigure[New-class average accuracy.]{
    \includegraphics[width=0.32\textwidth]{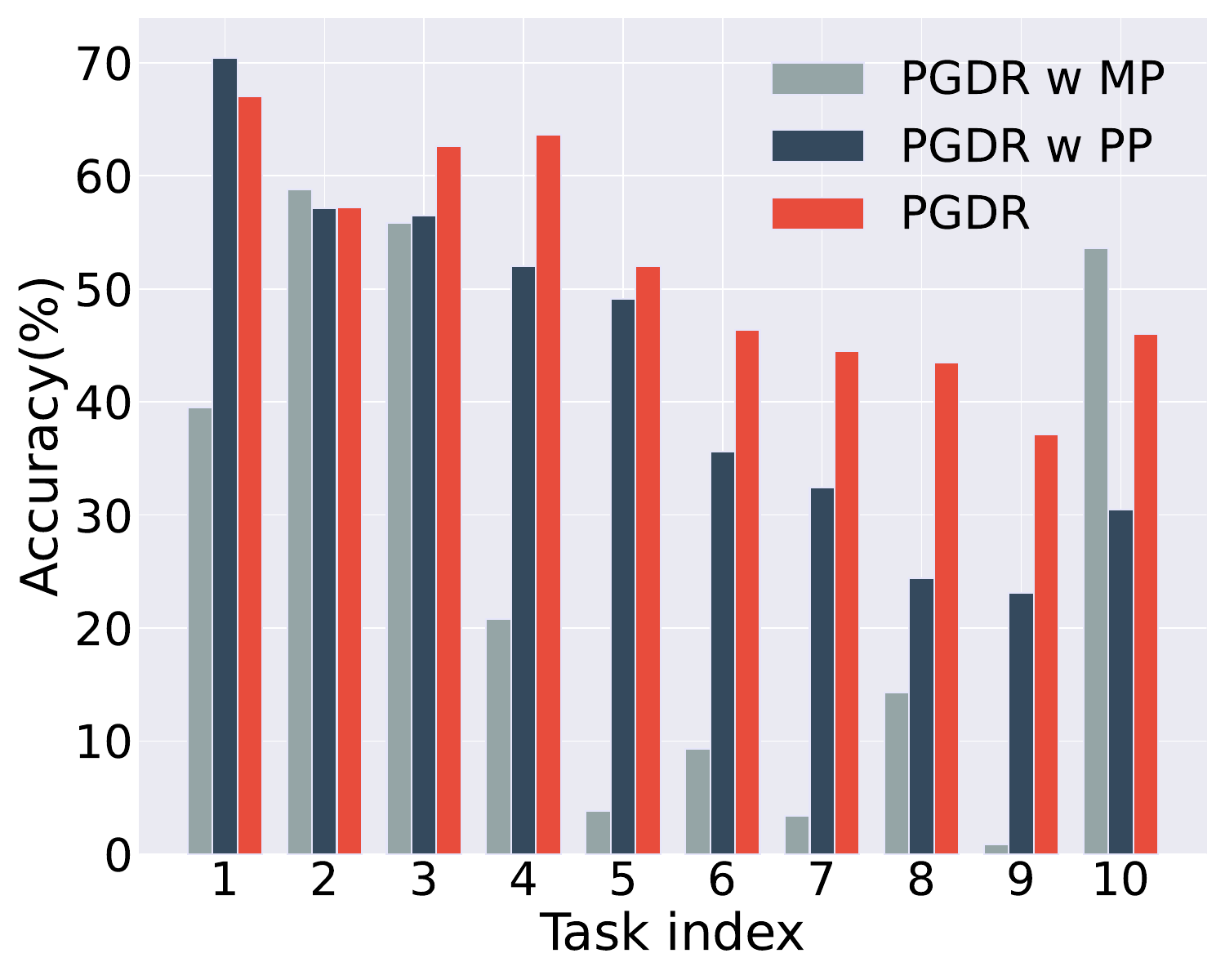}
    \label{fig:Disambiguation_new}
  }
  \hfill
  \subfigure[Old-class average accuracy.]{
    \includegraphics[width=0.32\textwidth]{source/Disambiguation_old.pdf}
    \label{fig:Disambiguation_old}
  }
  \caption{Ablation experiment about disambiguation module on Tiny-ImageNet ($q$=0.2).}
  \label{fig:fig5}
\end{figure*}

\begin{figure*}[h!]
  \centering
  \subfigure[All-class average accuracy.]{
    \includegraphics[width=0.32\textwidth]{source/memory_all.pdf}
    \label{fig:mem_all}
  }
  \hfill
  \subfigure[New-class average accuracy.]{
    \includegraphics[width=0.32\textwidth]{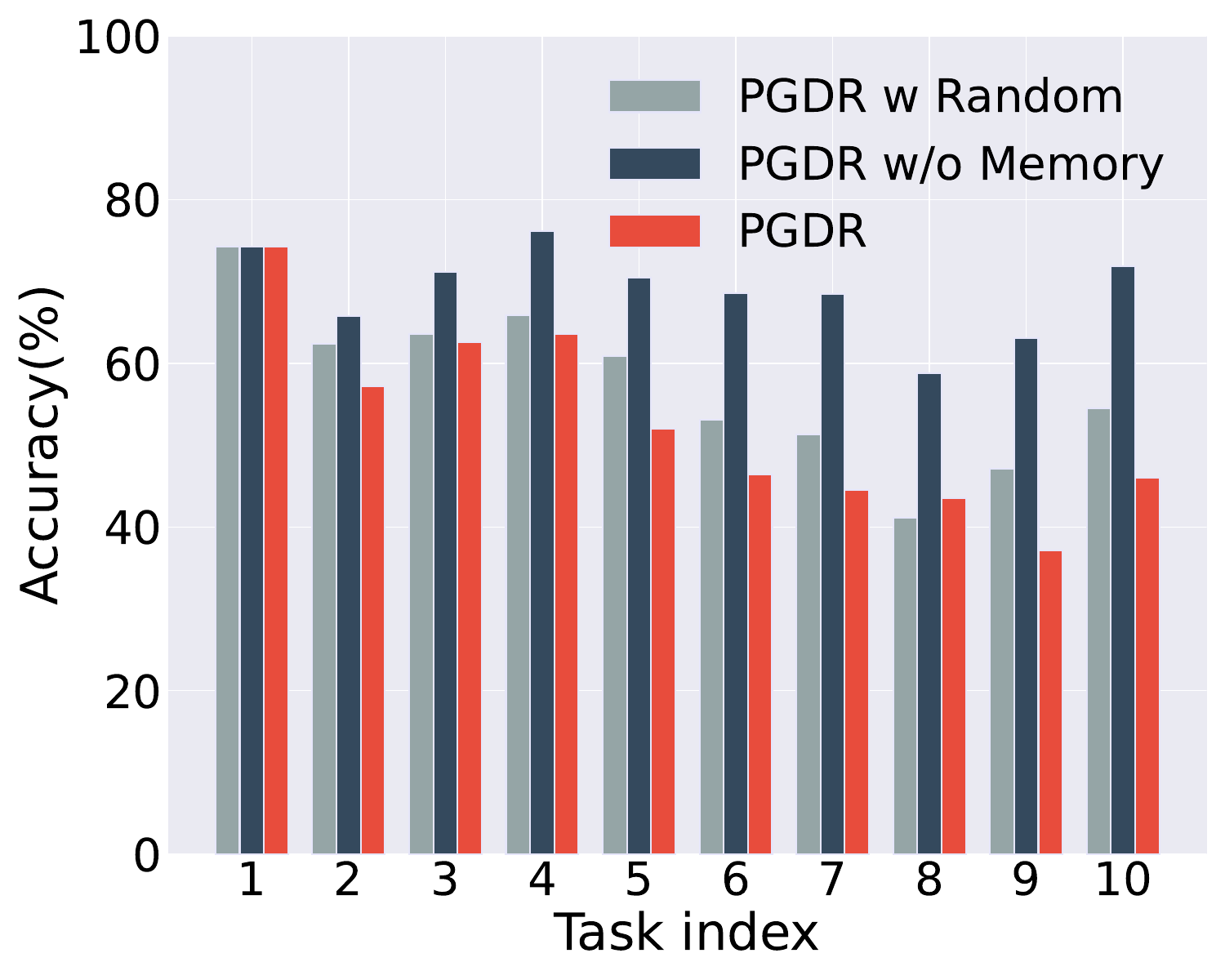}
    \label{fig:mem_new}
  }
  \hfill
  \subfigure[Old-class average accuracy.]{
    \includegraphics[width=0.32\textwidth]{source/memory_old.pdf}
    \label{fig:mem_old}
  }
  \caption{Ablation experiment about memory replay module on Tiny-ImageNet ($q$=0.2).}
  \label{fig:fig6}
\end{figure*}

\begin{table*}\centering
\begin{tabular}{c|c|cccc}
\toprule
\multirow{2}{*}{\textbf{Dataset}}  & \multirow{2}{*}{\textbf{Method}} & \multicolumn{2}{c}{$q=0.1$} & \multicolumn{2}{c}{$q=0.2$} \\\cmidrule(lr){3-4}\cmidrule(lr){5-6}%
                          &                         & 10-blurry   & 30-blurry   & 10-blurry   & 30-blurry   \\
\midrule
\multirow{2}{*}{CIFAR100} 
                          &  \cellcolor{gray!30} PGDR   & \cellcolor{gray!30}\textbf{64.40}	& \cellcolor{gray!30}\textbf{65.54}	& \cellcolor{gray!30}\textbf{63.83}	& \cellcolor{gray!30}\textbf{64.58} \\
                          & PGDR w Linear Classifier & 59.30 & 61.15  & 57.84 & 60.23   \\
\midrule
\multirow{2}{*}{Tiny-ImageNet}  
                          &   \cellcolor{gray!30}PGDR   &\cellcolor{gray!30}\textbf{41.60}	&\cellcolor{gray!30}\textbf{42.39}	&\cellcolor{gray!30}\textbf{38.46}	&\cellcolor{gray!30}\textbf{39.83}  \\
                          & PGDR w Linear Classifier  & 32.85 & 35.45 & 31.78 & 33.35 \\
\bottomrule
\end{tabular}
\caption{Performance comparisons of PGDR and \textit{PGDR w Linear Classifier} on CIFAR100 and Tiny-ImageNet.}\label{tab:classification}
\vspace{0.5cm}

\begin{tabular}{@{}c|cccc@{}}
\toprule
\textbf{Ablation} & \textbf{Disambiguation} & \textbf{Memory Replay} & \textbf{10-blurry} & \textbf{30-blurry}\\
\midrule
\rowcolor[gray]{.85} PGDR &\checkmark &\checkmark & \textbf {68.49} & \textbf {70.85} \\
PGDR w MP & Model Prediction  &\checkmark  & 67.90 & 68.36\\
PGDR w PP & Prototype Prediction  &\checkmark  &30.73  &27.05  \\\hline
PGDR w/o Memory &$\checkmark$ &$\times$ &26.65 & 26.64  \\
PGDR w Random &\checkmark & Random & 63.03  & 65.99 \\
PGDR w Distance &\checkmark & Prototype Distance & 67.15  & 68.34 \\
\bottomrule
\end{tabular}
\caption{Ablation study of PGDR with contrastive learning on CIFAR100 with $q=0.2$ at IPLL 10 tasks.}\label{tab:pgdr_cl}
\end{table*}

\subsection{More Ablation}
In this section, we still exclude the contrastive learning and the Kullback-Leibler divergence to analyze the performance of PGDR independently. Additionally, we conduct an analysis of PGDR's performance when incorporating contrastive learning.

\paragraph{Effects of the label-updating parameter $\beta$.} 
We analyze the parameter $\beta$ used in the label updating based on the prediction of the model. Table \ref{tab:beta} shows the performance of PGDR with varying $\beta$ on CIFAR100. When the $\beta$ is too high, pseudo-labels are primarily composed of disambiguated labels obtained in the initial stage. It becomes challenging to further disambiguate pseudo-labels using model outputs, leading to a decline in performance. Meanwhile, within a broader range of $\beta$ variations, the changes in the model's classification performance are relatively small, indicating the robustness of our approach.

\paragraph{Effects of disambiguation threshold $\alpha$.}
We analyze the confidence threshold $\alpha$ used in the label disambiguation module to distinguish between new and old classes. Table~\ref{tab:alpha} shows the performance of PGDR with varying $\alpha$ on CIFAR100. It can be observed that as the threshold increases, there is a slight fluctuation in accuracy. This reflects the robust characteristics of the model towards thresholds $\alpha$ in our solution, demonstrating its strong adaptability to dynamic environments.

\paragraph{Effects of selection parameters $K$ and $N_d$.} We analyze the number $K$ of nearest neighbor samples and the storage quantity $N_d$ of diverse samples within the data replay module. As $N_d$ is determined by the product of the single-class sample storage limit and the preset ratio, we analyze the preset ratio corresponding to $N_d$. In Figure~\ref{fig:k}, it is evident that inadequate selection of neighboring samples results in difficulty representing local samples, consequently leading to the ineffectiveness of the diversity strategy. Conversely, when the number of selected neighboring samples is excessively high, it results in a limited diversity of filtered samples, hindering their effectiveness. In Figure~\ref{fig:n_d}, it can be observed that a limited quantity of diverse samples in storage is unfavorable for adequately covering the sample space, resulting in a decrease in accuracy. When storing an excessive number of diverse samples, it reduces the proportion of representative samples in storage, impacting the availability of stored data.

\paragraph{The impact of storage limitation $m$.} To analyze the impact of storage limits on mitigating the effects of category-aware perception balance, we set the storage limit $m$ at 500, 1000, 2000, 3000, and 4000. The experiment is conducted on 10-blurry in CIFAR100 with a flipping probability $q$=0.2. As shown in Figure~\ref{fig:m}, with the arrival of new data, when storage capacity is low, there is a catastrophic decline in performance, clearly demonstrating that increasing the storage capacity effectively alleviates catastrophic forgetting. It assists the model in recognizing old class patterns and mitigates bias towards new classes.

\begin{figure}
  \centering
  \includegraphics[width=0.8\linewidth]{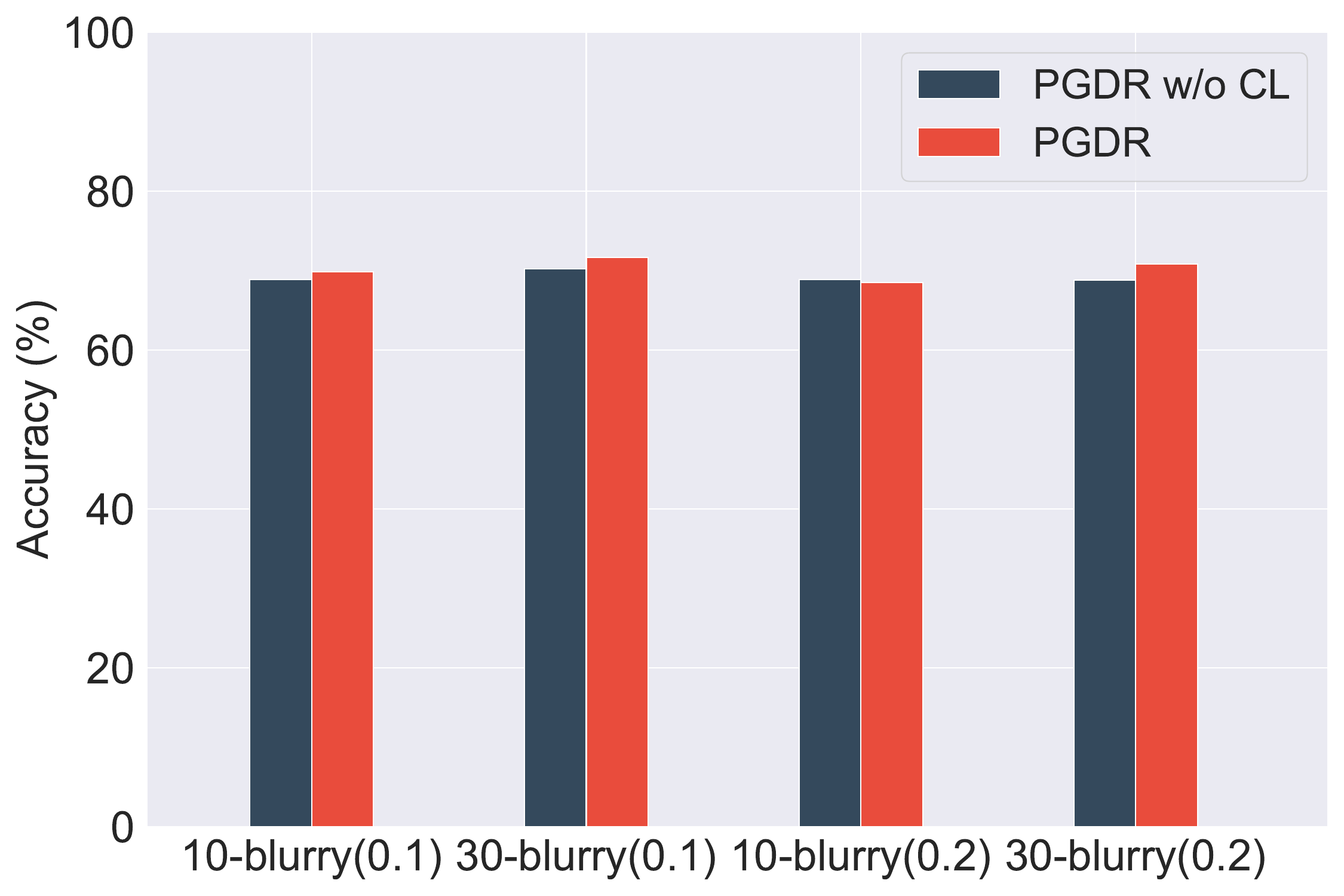}
   \caption{Ablation experiment of the contrastive learning module on CIFAR100. The sets include 10-blurry ($q=0.1$ and $q=0.2$) and 30-blurry ($q=0.1$ and $q=0.2$).}
   \label{fig:w_o_CL}
\end{figure}

\paragraph{The impact of $\mathcal{L}_\text{kd}$ and $\mathcal{L}_\text{cr}$.}
We analyze the contributions of consistency regularization (CR) and knowledge distillation (KD) within our framework. Specifically, we compare PGDR with two variants: 1) \textit{PGDR w/o CR} removes consistency regularization; 2) \textit{PGDR w/o KD} removes knowledge distillation. In Figure~\ref{fig:cr_kd}, it is evident that \textit{PGDR w/o CR} performs worse than both PGDR and \textit{PGDR w/o KD} in each task. This reflects the high reliability of our disambiguation strategy, which can provide accurate information for consistency regularization to facilitate the learning of ambiguous samples. The effectiveness of \textit{PGDR w/o KD} did not experience a significant decline with the arrival of new data, which demonstrates the excellent performance of our replay strategy in mitigating forgetting.

\paragraph{Supplementary ablation results on modules.} In the previous sections, we present partial results about the disambiguation module and the memory replay module. The detailed comparison results for each task on the Tiny-ImageNet ($q$=0.2) are illustrated in Figure \ref{fig:fig5} and Figure \ref{fig:fig6}. For the disambiguation module, Figure \ref{fig:fig5} shows that PGDR significantly outperforms \textit{PGDR w MP} and \textit{PGDR w PP} on new categories. Regarding the memory replay module, although \textit{PGDR w/o Memory} achieves favorable results on new categories, it suffers from severe forgetting of old category. Our proposed approach effectively balances learning between new and old categories, resulting in excellent overall performance.

\paragraph{Effect of evaluation strategy.} \label{evaluation_strategy}
To verify the classification effect of the feature prototype classifier, we remove the feature prototype classifier and classify using a linear classifier (\textit{PGDR w Linear Classifier}) in the test phase. As shown in Table~\ref{tab:classification}, the PGDR effect is much higher than the \textit{PGDR w Linear Classifier}. Using a linear classifier leads to significant classification bias, while using the slower updating feature prototype classifier can retain sample knowledge better and achieve bias elimination in the testing phase.

\paragraph{Ablation of PGDR with PiCO’s contrastive learning.}
We conduct ablation experiments to verify the effectiveness of each component of PGDR with contrastive learning. Specifically, for the label disambiguation module, we choose the variants: 1) \textit{PGDR w MP} and 2) \textit{PGDR w PP}. To assess the effectiveness of the memory update strategy, we compare PGDR with three variants: 1) \textit{PGDR w/o Memory}, 2) \textit{PGDR w Random} and 3) \textit{PGDR w Distance}. As shown in Table~\ref{tab:pgdr_cl},  PGDR outperforms other variants, demonstrating the superiority of our method. Specifically, in the disambiguation module's ablation experiment, rhe effectiveness of \textit{PGDR w PP} is significantly reduced. This is because the addition of contrastive learning in our method intensifies the accumulation of errors in frequent prototype updates. In the experiment of memory update strategies, we outperform three variants, demonstrating that our memory update strategy remains effective even with the addition of contrastive learning. In addition, the performance of our method remains basically consistent before and after the removal of the contrastive learning module in the Figure~\ref{fig:w_o_CL}. This indicates that the CL module of PiCO is not the fundamental factor affecting PGDR performance, thus providing indirect validation of the feasibility of the proposed architecture.

\begin{figure}
  \centering
  \includegraphics[width=0.8\linewidth]{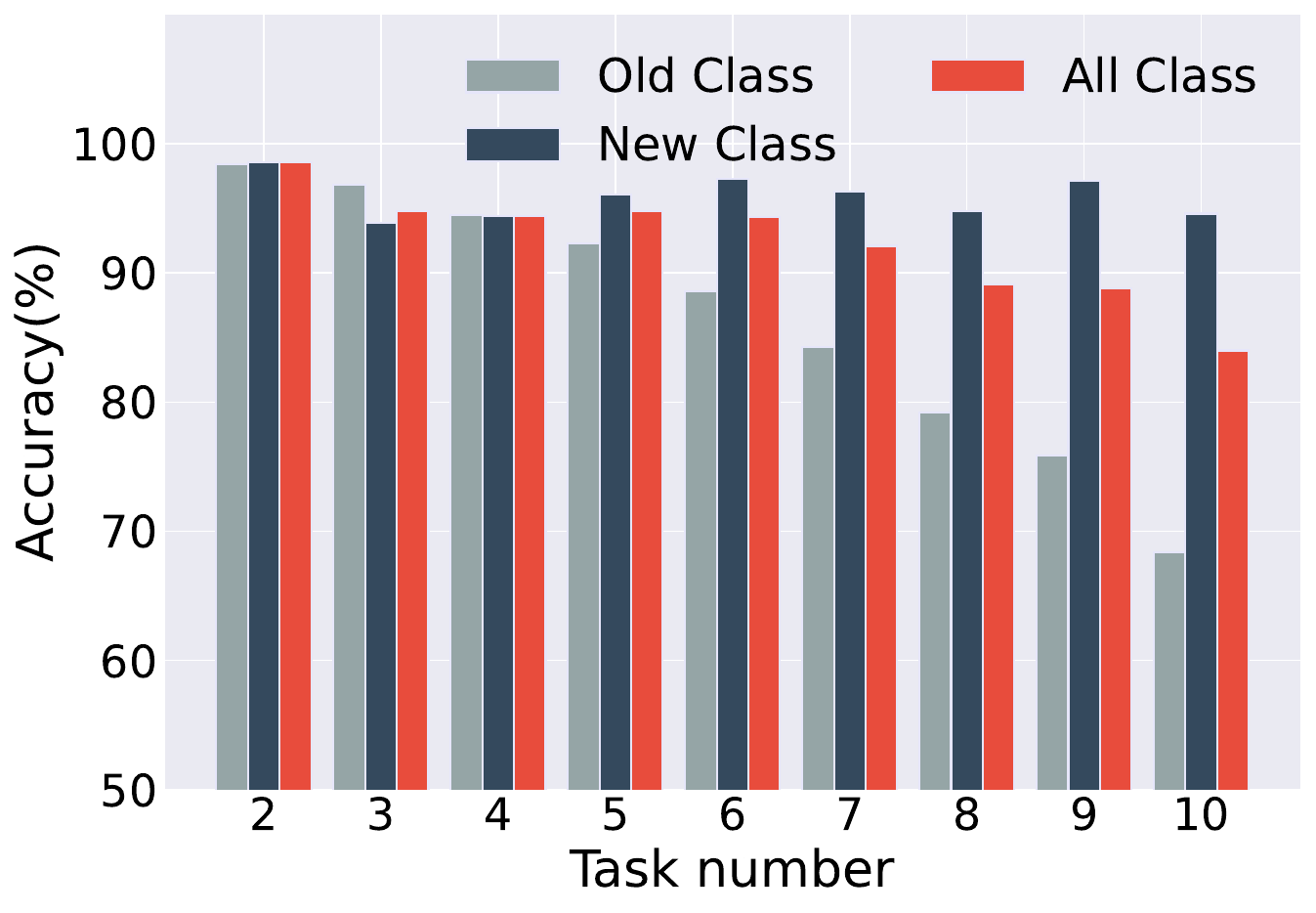}
   \caption{Performance of the prototype-based separation mechanism on CIFAR100.}
   \label{fig:separation}
\end{figure}

\begin{table}[!t]\centering
\scalebox{0.8}{
\begin{tabular}{c|cccc}
\toprule%
 \makebox[1.17cm][c]{\multirow{2}{*}{\textbf{Method}}} & \multicolumn{2}{c}{$q=0.1$} & \multicolumn{2}{c}{$q=0.2$} \\
\cmidrule(lr){2-3}\cmidrule(lr){4-5}
&10-blurry  &30-blurry  &10-blurry  &30-blurry \\
    \midrule
    PGDR & 69.84 & 71.68 & 68.49 & 70.85\\
    PGDR with Conf & 66.79 & 65.89 & 65.66 & 57.71\\
    PGDR with Entropy  & 66.67 & 65.73 & 66.02 & 56.27\\
\bottomrule%
\end{tabular}}
\caption{Ablation experiment of the prototype-based separation mechanism on CIFAR100.}\label{tab:separation}
\end{table}

\begin{table}
    \centering
    \begin{tabular}[t]{c|cc}
        \toprule
         & \textbf{CIFAR100} & \textbf{Tiny-ImageNet} \\
        \midrule
        iCaRL & 7.7 & 32.6 \\
        BiC & 9.0 & 35.3 \\
        WA & 7.8 & 33.6 \\
        ER-ACE & 7.6 & 31.6 \\
        ANCL & 9.8 & 39.5 \\
        PGDR & 8.5 & 34.2 \\
        \bottomrule
    \end{tabular}
    \caption{Running time (seconds/epoch) of each comparing approach on CIFAR100 and Tiny-ImageNet.}
    \label{tab:results}
\end{table}

\paragraph{The effect of the old/new data separation mechanism.} The mean accuracy of the separation mechanism is 92.32\% on CIFAR100. Specifically, 98.90\% of new-class samples are correctly detected as novel, and 86.49\% of true old-class samples are accurately identified. As shown in Figure~\ref{fig:separation}, accuracy declines with more tasks, with the most rapid decrease occurring in old-class samples due to catastrophic forgetting.
We also include comparisons with separation strategies such as calculating model confidence or uncertainty, to validate the effectiveness of prototype-based separation. As shown in Table~\ref{tab:separation}, our results demonstrate that as the degree of blur increases, both the confidence-based separation strategy and the entropy-based uncertainty method lead to a performance decline. We attribute this phenomenon to catastrophic forgetting, which causes the model to exhibit low confidence and high uncertainty for both the forgotten old knowledge and the previously unseen new knowledge, thereby significantly reducing the separation efficacy.

\section{Complexity Analysis}

Let $P$, $B$, $D$, and $K$ denote the dimension of the input sample, the batch size, the dimensionality of the feature, and the number of classes. Let $H$ denote the hidden dimensionalities of the model. Our time complexity is $O(B(PH+HD+DK))$. Below we show the running time of PGDR and baselines (seconds / epoch), where the components of PGDR show negligible effects on its efficiency. As shown in the Table~\ref{tab:results}, PGDR performs comparably to the baseline methods in terms of speed.

\section{Mean and Standard Deviation}
For each experiment, we conduct three independent runs, with the reported values in the main experimental table representing the mean results. The corresponding standard deviations are listed in Table~\ref{tab:std}.

\begin{table}[h]
    \centering
\begin{tabular}{l|cccc}
\toprule
\makebox[1.17cm][c]{\multirow{2}{*}{\textbf{Method}}} & \multicolumn{2}{c}{$q=0.1$} & \multicolumn{2}{c}{$q=0.2$} \\
\cmidrule(lr){2-3}\cmidrule(lr){4-5}
&10-blurry  &30-blurry  &10-blurry  &30-blurry \\
\midrule
 PiCO & 0.24 & 0.32 & 0.19 & 0.30 \\
 \midrule[0.01pt]
+iCaRL & 0.31 & 0.28 & 0.35 & 0.29 \\
+BiC  & 0.78 & 0.81 & 0.95 & 0.79 \\
+WA  & 0.41 & 0.52 & 0.38 & 0.36 \\
+ER-ACE  & 0.36 & 0.47 & 0.39 & 0.56 \\
+ANCL & 0.69 & 0.92 & 1.06 & 0.70 \\
+PGDR  & 0.88	& 0.89  & 0.94  & 0.78 \\
        \bottomrule
    \end{tabular}
    \caption{Standard deviation of methods on CIFAR100.}
    \label{tab:std}
\end{table}

\newpage
\section{Pseudo-code of PGDR}
\begin{algorithm}[htbp]
\renewcommand{\algorithmicrequire}{\textbf{Input:}}
\renewcommand{\algorithmicensure}{\textbf{Output:}}
\caption{Training Stage of Our Solution} \label{alg3} 
\begin{algorithmic}[1]
\REQUIRE { Dataset ${\cal D}_t$, memory ${\cal M}^t$,  model $f$, old model $f^{old}$, task $t$, prototypes $\bm{\mu}$} 
\ENSURE {Model parameter for $f$}
\IF{$t > 1$} 
\STATE{Get the distance $\bm e$ by Eq.(3)}
\STATE{Partition ${\cal D}_t$ into ${\cal D}_\text{old}$ and ${\cal D}_\text{new}$ }
\STATE{ Generate ${\cal S}'$ by Eq.(4)}\COMMENT{Re-allocate the candidate label sets}
\STATE{${ p_{i,j}} = \frac{1}{\lvert {\cal S}_i'\rvert} \mathbb{I} (j\in{\cal S}_i')$ }\COMMENT{Calculate the pseudo label}
\ENDIF
\FOR{ $epoch$=1,2,...}
\FOR{ $step$=1,2,...}
\STATE{${\bm{p}_i} \leftarrow \beta {\bm{p}_i} + (1 - \beta ){\bm{z}_i}$}\COMMENT{Update pseudo label}
\STATE{Minimize loss ${\mathcal{L}_\text{total}} = {\mathcal{L}_\text{ce}} + {\mathcal{L}_\text{kd}} + {\mathcal{L}_\text{cr}}$}
\ENDFOR
\ENDFOR
\STATE{${\bm{\mu} _c} = {\gamma {\bm{\mu} _c} + (1 - \gamma )\frac{1}{{|{\bm P_c}|}}\sum {{\bm P_c}} }$}\COMMENT{Update prototypes}
\STATE{Get the sum of distances to neighbors ${a_i}$}
\STATE{Update $\mathcal{M}_k^t$ by Eq.(9)}\COMMENT{Diverse samples}
\STATE{Update ${\cal M}^t_r$ by Eq.(7)}\COMMENT{Representative samples}
\STATE{${\cal M}^t={\cal M}^t_r \cup {\cal M}^t_k$}\COMMENT{Complete memory update}
\end{algorithmic} 
\end{algorithm}

\end{document}